# Knowledge Distillation in Vision Transformers: A Critical Review

Gousia Habib, Tausifa Jan Saleem, and Brejesh Lall
Bharti School of Telecommunication Technology and Management , Indian Institute of Technology Delhi, Delhi, India
gousiya.cstaff@iitd.ac.in, tausifa.cstaff@iitd.ac.in, Brejesh.lall@iitd.ac.in

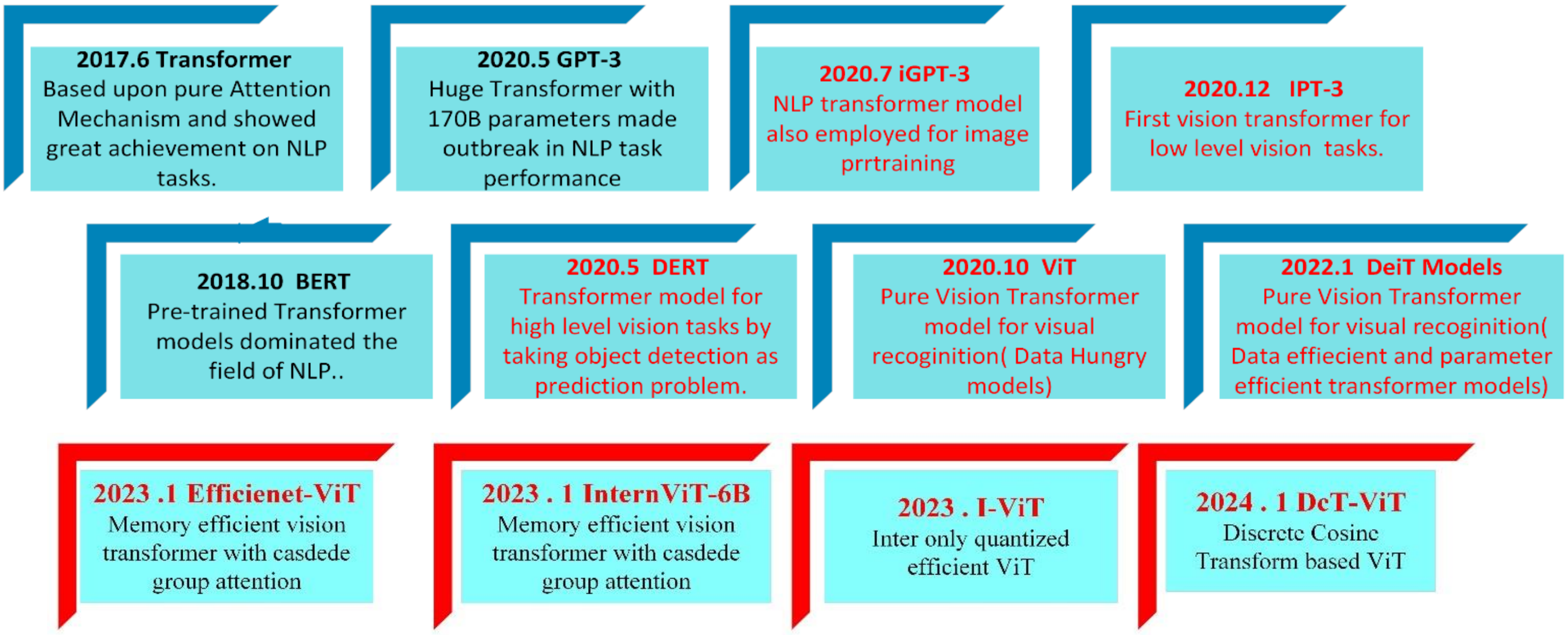

Transformer Milestones with red marked as Vision Transformers and others NLP.

**Abstract**: In Natural Language Processing (NLP), Transformers have already revolutionized the field by utilizing an attention-based encoder-decoder model. Recently, some pioneering works have employed Transformer-like architectures in Computer Vision (CV). In Response to such significant achievements, showing their effectiveness on three fundamental computer vision tasks such as classification, detection, and segmentation, as well as multiple sensory data streams (images, point clouds, and vision languages). Vision Transformers (ViTs) have demonstrated impressive performance improvements over Convolutional Neural Networks (CNNs) due to their competitive modelling capabilities. Demanding huge resources makes these models difficult to deploy on the resource constraint devices with strict latency requirements, which requires immediate research attention. Many solutions have been developed to combat the issues, such as compressive transformers and implementing compression functions such as dilated convolution, min-max pooling 1D convolution, etc. Model compression has recently attracted considerable research attention as a potential remedy. For such aggressive compression schemes, cutting-edge methods typically introduce complex compression pipelines. In addition, they rarely focus on smaller transformer models that have already been heavily compressed by Knowledge Distillation (KD) and need a systematic study to support their conclusions Which motivates us to review the techniques for effective compression of the transformers based on KD approach without compromising the performance. In this study, memory efficiency is not the primary objective, but a balance between computational efficiency, simplicity, and memory requirements is sought.

The paper reviews the aggressive KD based techniques for effective transformer compression (reducing these computational and memory requirements). The paper elucidates the role played by knowledge distillation in combating these challenges without compromising performance. Apart from KD methods for transformer compression. The paper also sheds some light on research challenges faced by the transformer that are open to address.

The entire paper is divided into three main sections; the first part presents a generic framework and introduction to the ViTs. The second part discusses the major challenges faced by ViTs in the CV field. Finally, the paper concludes with a comparative analysis of various compression methods implemented for ViTs to make them resource-efficient in terms of computational and memory requirements.

## 1) Introduction

In the evolving landscape of Artificial Intelligence (AI), Deep Neural Networks (DNNs) have established themselves as foundational elements, with Transformers emerging as an exemplary archetype for their ability to handle a wide range of tasks, including natural language processing and image recognition. Transformers uses a self-attention mechanism as the core innovation to weigh the importance of different parts of input data, thus capturing intricate patterns across large datasets. An integral part of Transformer models, self-attention enables direct interactions among all sequence elements, akin to relevance-based filtering. This characteristic enhances a model's contextual awareness, but exacerbates computational and memory usage in a quadratic manner, hindering scalability. Researchers have been developing efficient Transformers that maintain the model's effectiveness while minimizing memory footprints and computation demands in order to address these scalability issues. Knowledge Distillation (KD) has been crucial, offering tools for compressing models without compromising performance, making Transformers viable for on-device applications with limited RAM and processing power. In addition to enhancing their applicability, this nuanced approach leads to new avenues for their deployment in edge computing, mobile devices, and beyond, a significant step forward in making AI more accessible and useful for all. In devices with limited resources, the concept of model efficiency is often related to how much memory a model needs and how much computation it requires. Transformers are the subject of this discussion, especially their performance with complex data inputs and their ability to handle long sequences efficiently. A high level of efficiency is key for the technology to be applied to processing detailed content like text, images, and videos, since their broader integration and utility in advanced computing systems are heavily dependent on such efficiency. This paper presents a taxonomy representing milestones of efficient Transformer models, focusing on their innovations and applications both in natural language processing (NLP) and computer vision. Using vision Transformers, we examine the challenges of processing and understanding visual and text content, and we examine various distillation techniques that can be utilized to enhance Vision Transformers (ViT) performance, demonstrating that they are capable of handling a wide range of AI tasks.

In this survey, we focus on the dual aspects of efficiency and performance enhancement brought about by KD, emphasizing the need for models that combine computational demands with advanced capabilities, crucial for applications with limited resources.

## 2) Overview of Vision Transformers (ViT)

Through the use of self-attention mechanisms, Transformers have revolutionized both natural language processing and computer vision by allowing for a better understanding of sequences regardless of their length. In addition to playing a pivotal role in NLP, the architecture has inspired several variants, including Reformer and Performer, which boost memory and computational efficiency. Introducing the Vision Transformer (ViT) marks a significant advancement in image processing, surpassing the abilities of traditional convolutional neural networks and recurrent neural networks. In addition to enhancing the processing efficiency of Transformers, this advancement expands the range of applications, enabling them to serve a wide range of AI challenges across a variety of industries. Vanilla Transformer is a great model for overcoming the shortcomings of RNN models, but it still poses some critical challenges such as Limited Context Dependency and Context Fragmentation [1]. Transformer XL is an extension of the Transformer (it is extra-long). This Transformer extends the vanilla Transformer with a recurrence mechanism and relative positional encoding. For Transformer-XL, rather than computing the hidden state for each segment from scratch, it uses the hidden state from previously learned segments. In addition to solving the issues introduced in the vanilla transformer model, the model also overcomes the problem of long-term dependency. Keeping all these hidden states increases the computation and storage costs associated with attending each time step which is one of the major drawbacks of these revolutionary NLP models.

The simplicity and generalizability of transformer models have not made them a defacto standard in the NLP domain only. But transformers are also becoming popular in computer vision. Vision models based on Generative Adversarial Learning (GAN) models have been highly successful. But These models are usually built for highly specialized tasks. These models must be rebuilt or retrained whenever new tasks are introduced. In light of this, general-purpose learning models are becoming increasingly popular. Vision Transformer (ViT) is one example of such a model. Transformers originated from Natural Language Processing (NLP) applications [1], which aim to understand the text and draw meaningful conclusions. ViT is the application of Transformer in the image domain with only a few implementation modifications to accommodate the different data types. In particular, ViTs use different methods for tokenization and embedding. Despite this, the generic architecture remains the same. The input image is divided into image patches called visual tokens or tokens. With the encoded vector, the position of a patch in the image is embedded and fed into the transformer encoder network, which is essentially the same one that processes text input.

The generic methodology of the vision transformers is given in figure 1: As a result of their flexibility, ViTs are used across a wide range of vision tasks. A few examples of this are image classification [2], image-to-text/text-to-image generation [3], visual reasoning [4], association learning [5], and multi-modal learning [6]. Compared to state-of-the-art Convolutional Neural Networks (CNNs), the ViTs exhibit better performance over multiple benchmarks because of their competitive modelling capabilities. ViTs outperform the CNNs while requiring fewer computational resources for pretraining. When trained on enough data, ViT can surpass the performance of a similar state-of-the-art CNN using 4x fewer computational resources. Unlike CNN, self-attention in ViT enables the global embedding of information across the entire image. The model also learns how to encode the relative location of image patches to reconstruct the image's structure from training data. Vision transformer models are influenced by hyperparameters such as the optimizer, the network depth, and the parameters specific to the dataset.

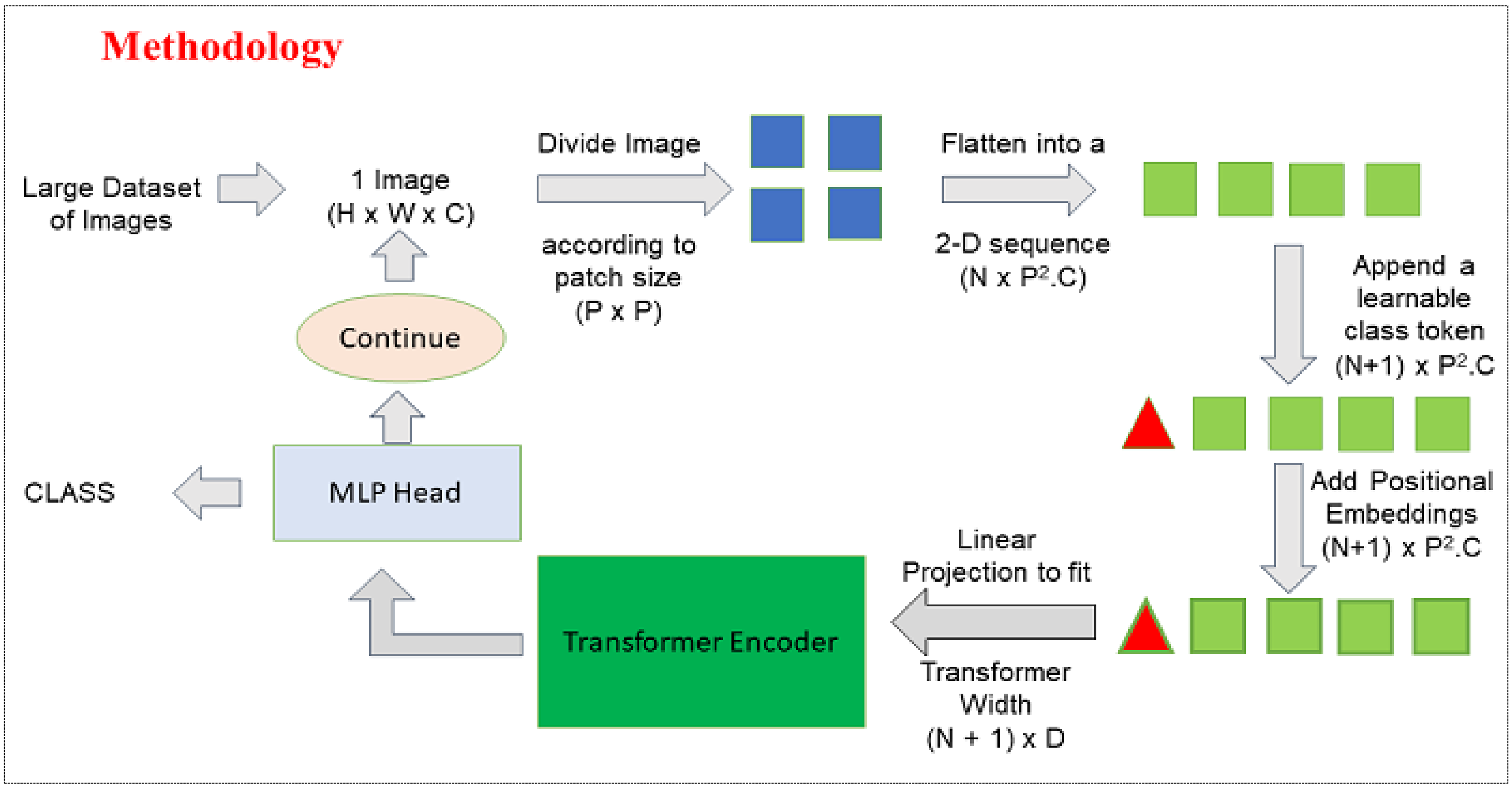

Figure 1. Generic working illustration of ViT

### 2.1) Open Research Challenges

The major challenges by ViT are summarized in the following proceeding subsections as:

**i) Positional Encoding**

The Transformer model is permutation equivariant. Position and segment embeddings are usually added to provide order and type information about the input tokens. Several works have explored how positions can be included in Transformers [6],[7],[8]. Relative position encodings are noted as being superior to absolute position encodings in many of [6],[7], and [8] studies. However, it is unclear what causes this difference. The authors in the paper [9] showed that Transformers with absolute position encodings are universal approximators of all sequence-to-sequence functions. This [9] demonstrates that absolute position encodings can capture positional information. Therefore, why do relative position encodings perform better than absolute position encodings? Therefore, the benefits and drawbacks of different position encoding methods need to be systematically studied and understood. It was hypothesized in [10] that the cross-correlation between word and position embeddings during attention computation might result in poor performance from absolute position encodings. Although cross terms are present in some relative position encoding methods [11],[12], these methods perform better than the other position encoding schemes. Various techniques are available for positional encoding. Simple example is given as:

$$PE_{(ps,j)} = \begin{cases} \sin\left(ps \cdot w_k\right) & \text{if} \quad j = 2k \\ \cos\left(ps \cdot w_k\right) & \text{if} \quad j = 2k+1 \end{cases} \tag{1}$$

where,

$$w_k = \left(\frac{1}{1000}\right)^{\frac{2k}{l}}, k = 1, 2, \ldots, \frac{l}{2} \tag{2}$$

$ps$ and $l$ are the position and the length of the vector, respectively, and $j$ is the index of each element within vector. Furthermore, a 2D interpolation complements pre-trained positional encoding to keep the patches in the same order even when the feeding images have arbitrary resolutions.

**ii) Weaker Inductive Bias**

While ViTs are excellent replacements for CNNs, one constraint that makes their application rather challenging is the requirement for large datasets. ViTs lack inductive biases inherent in CNNs, such as translation equivariance, which makes their generalization ability poor when trained on insufficient data. The situation changes if the models are trained on larger datasets such as Google Inhouse Large (JFT-300) containing 14M-300M images. The latter makes ViTs highly reliable on advanced data augmentation techniques such as Rand-Aug [13], Cut-Mix [14], Mix-up [15] etc. It has been found that CNNs can learn even if there is a relatively small amount of data, mainly due to their intrinsic inductive biases. In other words, inductive biases help the models learn faster and generalize better. Although convolutional architectures remain dominant in CV [16-18], there is still increased interest in their exploration and application.

In the wake of NLP success, several researchers conjugated and married CNNs with a self-attention mechanism [19,20]. In some research works, such as [21,22], authors replaced convolutions entirely with the self-attention mechanism. Despite their theoretical efficiency, these models must be successfully scaled on modern hardware accelerators, which leads to a suboptimal solution. Therefore, classic ResNet like architectures are still the gold standard in large-scale image recognition [23,24]. Alexey Dosovitskiy et al. [25] stated that although Transformers are a defacto standard in NLP, their application remains limited in vision tasks. Authors in this paper applied a standard Transformer architecture directly to images with the fewest modifications possible. They split an image into patches and then provided the linear embeddings of these patches as inputs to the Transformer. These patches acted as tokens, similar to the tokens in NLP. They trained the model in a supervised manner on image classification tasks using medium-sized datasets such as ImageNet [26], Cifar-100[27] and VTAB [28].

The models achieved a reasonable accuracy, although a few percentages points below ResNet. Authors observed that Training on large datasets trumps inductive bias. The authors concluded that ViTs could achieve excellent results when pre-trained on upstream tasks before being transferred to a downstream task. In multiple image recognition benchmarks, ViTs perform similarly to the state-of-the-art or even beat it in some situations when pre-trained on large ImageNet-21k and JFT-300M datasets. Specifically, the best model achieves 88.55% accuracy on ImageNet, 90.72% on ImageNet-Real, 94.55% on CIFAR-100, and 77.63% on VTAB. The proposed model in the paper [19] is similar to the model given by Cordonnier et al. [29], in which the patches of small size $2 \times 2$ is extracted from an input image, and the full self-attention mechanism is applied for the classification task. In contrast to the model proposed by Cordonnier et al. [29], Alexey Dosovitskiy et al. [25] demonstrated that large-scale pretraining could make ViTs competitive with (or even better than) state-of-the-art CNNs. Cordonnier et al. [29] used a small patch size of $2 \times 2$ pixels, making their model apply only to small-resolution images.

In contrast, Alexey Dosovitskiy [25] used the patch size of $14 \times 14$ and $16 \times 16$ and, hence, could handle the medium-resolution images. Chen et al. [30] proposed the image GPT (iGPT) model. The Transformers were applied to image pixels after reducing image resolution and color space rather than directly using image patches. The model was trained as an unsupervised generative model and then probed linearly to achieve a maximum classification accuracy of 72% on ImageNet. Alexey Dosovitskiy et al. [25] faced a major challenge while finetuning the proposed model for high-resolution images. To overcome the challenge, instead of using pre-trained prediction heads, they added $D \times K$ zero-initialized feedforward layers ($K$ represents the number of classes in the downstream task, and $D$ represents the model depth).

Touvron et al. [31] and Kolesnikov et al. [32] suggested that it is more advantageous to finetune at a higher resolution than to pretrain the model from scratch. They observed that feeding the inputs of higher resolution to the proposed model (keeping the patch size the same) leads to larger effective sequence lengths. Although, ViTs support sequences up to arbitrary length (within memory constraints), pre-trained embeddings may no longer be useful in this case. As a result, Touvron et al. [31] and Kolesnikov et al. [32] performed 2D interpolation using the original image location of the pre-trained position embeddings. As far as the ViTs are concerned, this resolution adjustment and patch extraction are the only points where a bias is manually injected. The architecture proposed by Alexey Dosovitskiy et al. [25] is given in figure 2:

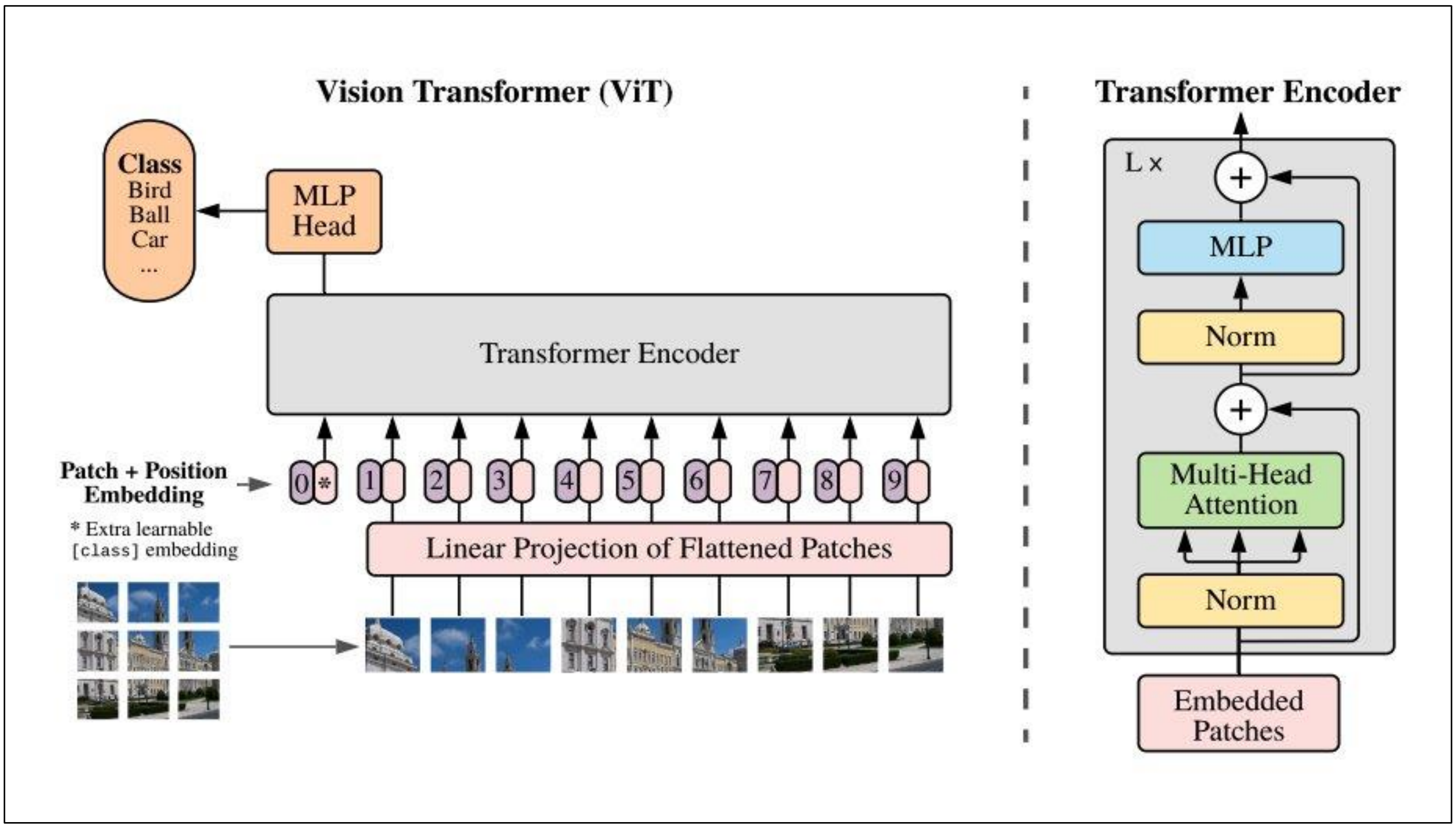

Figure 2. The generic architecture of ViT

### iii) Quadratic Complexity of Attention Mechanism

ViT implements scaling of dot-product attention, which is similar to the general attention mechanism as in case NLP transformers. For each query, $q$, the scaled dot-product attention first computes a dot product with all the keys, $k$. This is followed by a soft-max function which divides each result by the number of keys, $\sqrt{d_k}$ Consequently, it obtains the weights used to scale the values, $v$. As a result, the computations performed by scaled dot-product attention can be applied efficiently to the entire pool of queries at the same time. As inputs to the attention function, Q, K, and V are used as follows:

$$\text{Attention}(Q,K,V) = \text{Softmax}\left(\frac{QK^T}{\sqrt{d_k}}\right) \quad (3)$$

The multi-head attention mechanism linearly projects queries, keys, and values h times, using a different learned projection every time. To produce a final result, the single attention mechanism is applied to each of these h projections in parallel, producing h outputs, which are then concatenated and projected again. A multi-head attention model is capable of attending to information from multiple representation subspaces from multiple positions simultaneously. In the case of a single attention head, averaging prevents this from happening. The general equation for the MHSA is given as:

$$\left.\begin{aligned} &\mathrm{MHSA}(Q,K,V) = Concat(head_1,\cdots,head_h)W^O \\ &\text{where} \qquad head_i = \mathrm{Attention}\left(QW_i^Q, KW_i^K, VW_i^V\right) \end{aligned}\right\} \qquad (4)$$

Parameter matrices as $W_i^Q \in \mathbf{R}^{d_{\text{model}} \times d_q}$, $W_i^K \in \mathbf{R}^{d_{\text{model}} \times d_k}$ and $W_i^V \in \mathbf{R}^{d_{\text{model}} \times d_v}$ are actually the linear projections of $Q$, $K$ and $V$ with $W^o \in \mathbf{R}^{hd_v \times d_{\text{model}}}$ is the initial trainable matrix jointly trained with model. The attention mechanism of ViT has quadratic complexity requiring $O(n^2)$ memory and compute capability, which makes these models possess billions of parameters and are, therefore, both resource-intensive and computationally complex. Demanding huge resources makes these models difficult to deploy on the resource constraint devices with strict latency requirements, which requires immediate research attention and motivates us to review the techniques for effective compression of the transformers without compromising the performance.

Knowledge Distillation (KD) comes up with multiple advantages as one stop solution, focusing on solving inductive bias, positional encoding, and quadratic complexity problems beyond model compression. Through KD, complex models can be transformed into simpler, more efficient counterparts that improve learning efficiency and model understanding without requiring a high amount of computational power. Through this process, positional encoding issues can be mitigated by simplifying sequence representations, weaker inductive bias can be mitigated by transferring more learning patterns, and quadratic complexity can be addressed by streamlining model operations, thereby significantly reducing computational demands and memory footprints in addition to above mentioned critical challenges. This study aims at optimizing Transformer models holistically, not just on memory efficiency, but also on an equal balance between computational speed, model simplicity, and memory consumption. It represents a landmark contribution to the field by compiling and analyzing developments in Vision Transformers (ViTs) through Knowledge Distillation techniques.

## 3) Preliminaries: KD

Many solutions have been developed to combat the issues, such as compressive transformers and implementing compression functions such as dilated convolution, min-max pooling 1D convolution, etc. Model compression has recently attracted considerable research attention as a potential remedy. It has been proposed to fit large NLP models on resource-constrained devices using extreme compression, particularly ultra-low bit precision (binary/ternary) quantization [33]. For such aggressive compression schemes, cutting-edge methods typically introduce complex compression pipelines, e.g., multiple stages of expensive knowledge distillation with extensive hyperparameter tuning. In addition, they rarely focus on smaller transformer models that have already been heavily compressed by knowledge distillation and need a systematic study to support their conclusions. As a result of Knowledge Distillation, a smaller model (known as the student) can be trained using the outputs (from various intermediate functional components) of a larger model (known as the teacher). There are times when information is provided through an intermediary model (commonly known as a teaching assistant) [34],[35]. A knowledge distillation consists of three major components: knowledge, distilling algorithm, and teacher-student architecture. Framework for knowledge-sharing between teachers and students. The framework for knowledge-sharing between teachers and students is given in figure.1, which gives an overview of the knowledge distillation process and its relationship with the adjacent sections. The illustration of all the segments can be best understood from figure 1 given as:

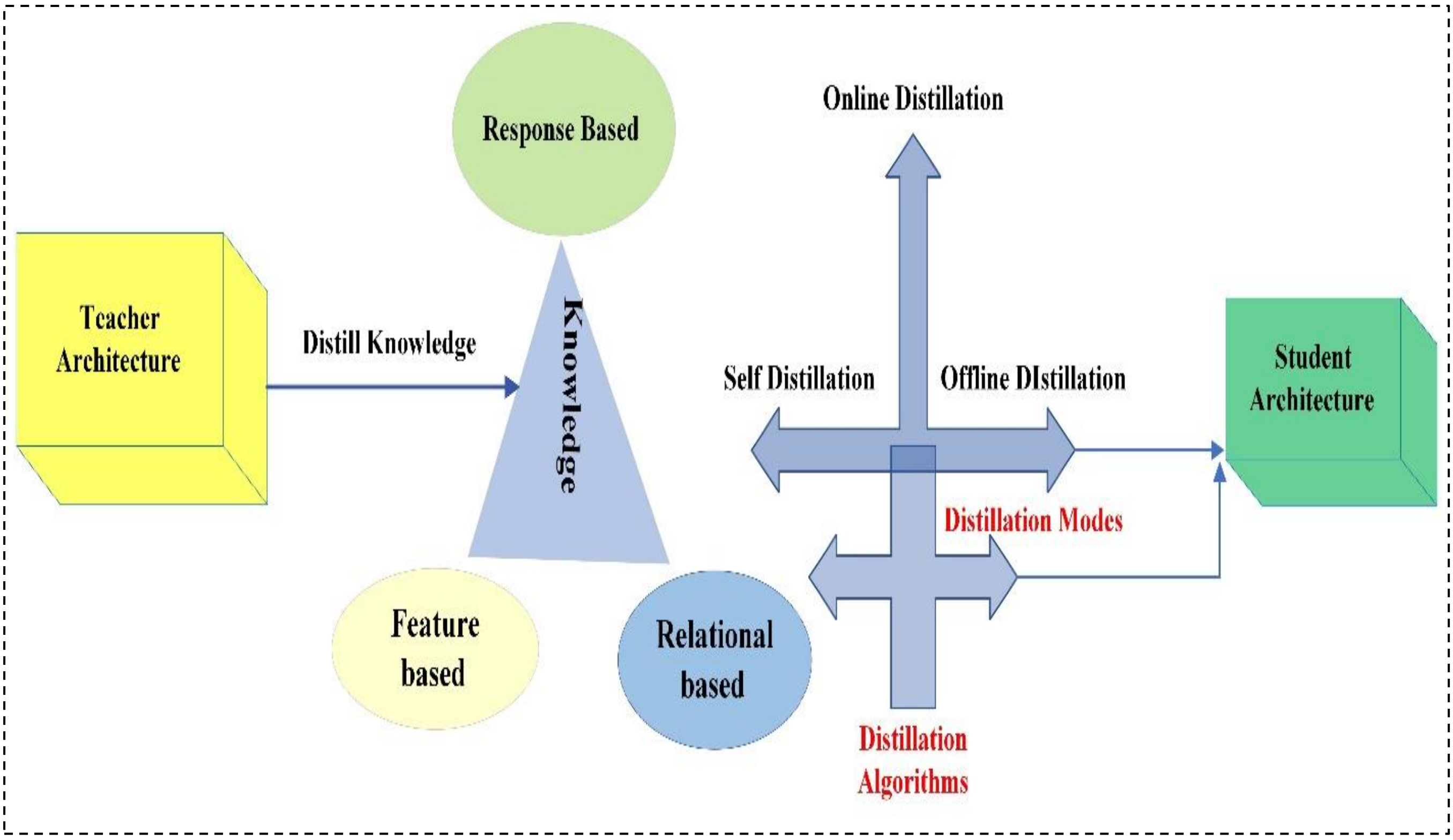

Fig.3. Universal Knowledge Distillation frame.

It is inferred from the above discussion that ViTs need better generalization when trained on insufficient amounts of data. Enormous training images must be used to achieve excellent performance. Therefore, a data-efficient transformer solution is urgently needed. Training on small datasets with ViT increases dependence on regularization or data augmentation techniques. If trained on massive datasets, ViTs can achieve competitive performance despite lacking intrinsic properties of convolutions. However, due to the stacking of self-attention modules on multiple heads, ViTs remain prohibitively expensive. ViT compression has only recently come into existence, and the prior works mainly focus on a few aspects, in contrast to the extensive literature and remarkable success of CNN compression.

This paper contains more than eighty research papers exploring the research works about ViT compression. Among the most popular methods for ViT compression to reduce the computational cost is Knowledge Distillation (KD) [36], which enables effective knowledge transfer between a large teacher model and a smaller student network. According to Gou et al. [37], there are three types of knowledge to be distilled (see Figure. 3 a-c): response-based knowledge [38], [39], [40], feature-based knowledge [41],[42], and relation-based knowledge [43], [44], [45]. Response-based and feature-based knowledge distillation uses the output from one specific layer of the teacher model (the last or intermediate layer, respectively). Still, relation-based distillation uses the output from more than one layer simultaneously. There is a predominant focus on CNN-to-CNN transfer in these paradigms [38], [39], [41]. Still, they need to sufficiently cover KD from transformer-specific intermediate blocks, such as patch embeddings, resulting in less satisfactory results. Given the high computational costs of ViTs and the extensive pretraining they require, the question arises: could lightweight models be created using different KD methods?

To answer this question or to achieve our goal, we discussed various solutions that help us to distil the knowledge from the heavy teacher model to the lightweight student model under various distillation schemes. While lighter than the teacher model, the student model has the same expressive power as the teacher model with significantly fewer parameters. KD in computer vision is divided into three main categories according to Guo et al. [37] response-based knowledge, relation-based knowledge and feature-based knowledge distillation method

**3.1) Loss functions and theoretically Investigating the role of temperature $\upsilon$ softening hyperparameter**

**Kullback-Leibler divergence (KL Divergence)**: It describes how one probability distribution differs from another. According to Bayesian theory, there is some true distribution $P(X)$ we would like to estimate it with an approximate distribution $Q(X)$. Essentially, the KL divergence calculates the distance between an approximate distribution Q and a true distribution P. In mathematics, consider two probability distributions on some space *X*, *P* and *Q*. *KL* Divergence from *Q* to *P* can be written as $D_{kl}(P \| Q)$

$$D_{kl}(P \| Q) = E_x \sim P\left[\log \frac{P(X)}{Q(X)}\right] \tag{6}$$

We would like to point out a few immediate observations about this definition. *KL* Divergence is not symmetrical i, e $D_{kl}(P \| Q) \neq D_{kl}(Q \| P)$. Therefore, it cannot be considered a distance metric. *KL* Divergence takes values in the interval defined as $[0, \infty)$ in case when *P* and *Q* are same distributions. For finiteness of *KL* Divergence, it is essential for the support of *P* needs to be contained in the support of *Q*.
For a network F, the j-th value of the softened probability vector with a temperature scaling hyperparameter $\upsilon$ is given by $p_j^F(\upsilon) = \exp\left(w_j^F/\upsilon\right) \Big/ \sum_j \exp\left(w_j^F/\upsilon\right)$, where $w_j^F$ is the j-th value of the logit vector $w^F$ for a sample y, than for the student network the loss function $\Lambda$ is convex combination of cross-entropy loss $\Lambda_{ce}$ Kullback Lieber divergence loss $\Lambda_{kl}$ that is

$$\Lambda = \lambda_1 \Lambda_{ce}\left(p^s(1), y\right) + \lambda_2 \Lambda_{kl}\left(p^s(\upsilon), p^t(\upsilon)\right) \tag{7}$$

Where, $\lambda_1 + \lambda_2 = 1$ and

$$\Lambda_{ce}\left(p^s(1), y\right) = \sum_j -y_j \log p_j^s(1) \ , \ \Lambda_{kl}\left(p^s(\upsilon), p^t(\upsilon)\right) = \upsilon^2 \sum_j p_j^t(\upsilon) \log\left(\frac{p_j^t(\upsilon)}{p_j^s(\upsilon)}\right)$$

Note that here's indicates the student network, t indicates the teacher network, y is a one-hot label vector of a sample x and $\lambda_1, \lambda_2$ are the parameters.

In [91] it was shown given a single sample x, the gradient of $\Lambda_{kl}$ with respect to $w_j^s$ is given by

$$\frac{\partial \Lambda_{kl}}{\partial w_j^s} = \upsilon\left(p_j^s(\upsilon) - p_j^t(\upsilon)\right) \tag{8}$$

**Mean Square Error (MSE)**: For large value of $\upsilon \to \infty$, the term $w_j^s/\upsilon$ becomes quite small, hence $\exp\left(w_j^F/\upsilon\right) \approx 1 + w_j^F/\upsilon$, so the gradient can be written as

$$\frac{\partial \Lambda_{kl}}{\partial w_j^s} = \upsilon\left(\frac{1 + w_j^s/\upsilon}{K + \sum_j w_j^s/\upsilon} - \frac{1 + w_j^t/\upsilon}{K + \sum_j w_j^t/\upsilon}\right) \tag{9}$$

Assuming zero-mean teacher and student logit, that is $\sum_j w_j^s = 0$ and $\sum_j w_j^t = 0$ than $\frac{\partial \Lambda_{kl}}{\partial w_j^s} = \frac{1}{K}\left(w_j^s - w_j^t\right)$. This indicates that minimizing $\Lambda_{kl}$ is equivalent to minimizing the mean squared error $\Lambda_{mse}$, that is, $\left\|w^s - w^t\right\|_2^2$, under a sufficiently large temperature $\upsilon$ and the zero-mean logit assumption for both the teacher and the student.

**Cosine Similarity Loss**: The cosine similarity between two vectors is defined mathematically as the product of the Euclidean norm or magnitude of each vector divided by its dot product. Where Similarity is given as:

$$\text{cosine similarity} = \cos(\theta) = \frac{\mathbf{x} \cdot \mathbf{y}}{\|\mathbf{x}\|\|\mathbf{y}\|} \qquad (10)$$

**Cross-Entropy Loss**: Calculates how closely the predicted value is to the actual value using cross-entropy. Mathematically the loss term is given as:

$$\text{loss}(A, B) = -\sum A \log(B) \qquad (11)$$

### 4) Different KD approaches for effective ViT compression

This section discusses the various solutions for aggressive compression of ViTs using different knowledge distillation approaches. Knowledge Distillation (KD) effectively improves the performance of lightweight student networks because teachers can directly impart domain knowledge to students. A large model pre-trained on large datasets is often found to achieve better results than a small model, as small models are susceptible to saturation (or underfitting) when their data scale is increased [46]. Small models can achieve the potential of large-scale pretraining data by distilling large-scale pretraining data and using a powerful model as the teacher. In the meantime, the distilled small models can be easily applied to downstream tasks since they have learned much about generalizing from both the large and large-scale pretraining datasets.

Further, recent advances in knowledge distillation (KD) techniques have enabled the development of hybrid approaches that combine the strengths of different KD methodologies to further enhance the compression and performance of Vision Transformers (ViTs) in resource-constrained environments. Students will benefit from learning paths that combine feature-based, relation-based, and logits-based distillation strategies. In order to provide the student with richer and more nuanced guidance, these hybrid methods integrate multiple aspects of the teacher model's knowledge, such as feature representations, inter-layer relationships, and output predictions. Thus, the student models not only closely mimic the teacher's output, but also inherit a deeper understanding of the data representations and relationships, which ultimately leads to improved generalization. Also, some approaches incorporate attention mechanisms to selectively focus on the most informative parts of the teacher's knowledge, optimizing the distillation process. With the continued development of hybrid KD technologies, they offer promising avenues for deploying highly efficient yet powerful ViTs across a variety of applications where resources are limited. The following presents some of the solutions proposed in the literature to deploy ViTs in resource-constrained environments using the KD approach.

The overall taxonomy of KD approaches discussed in this paper are given in figure 4.

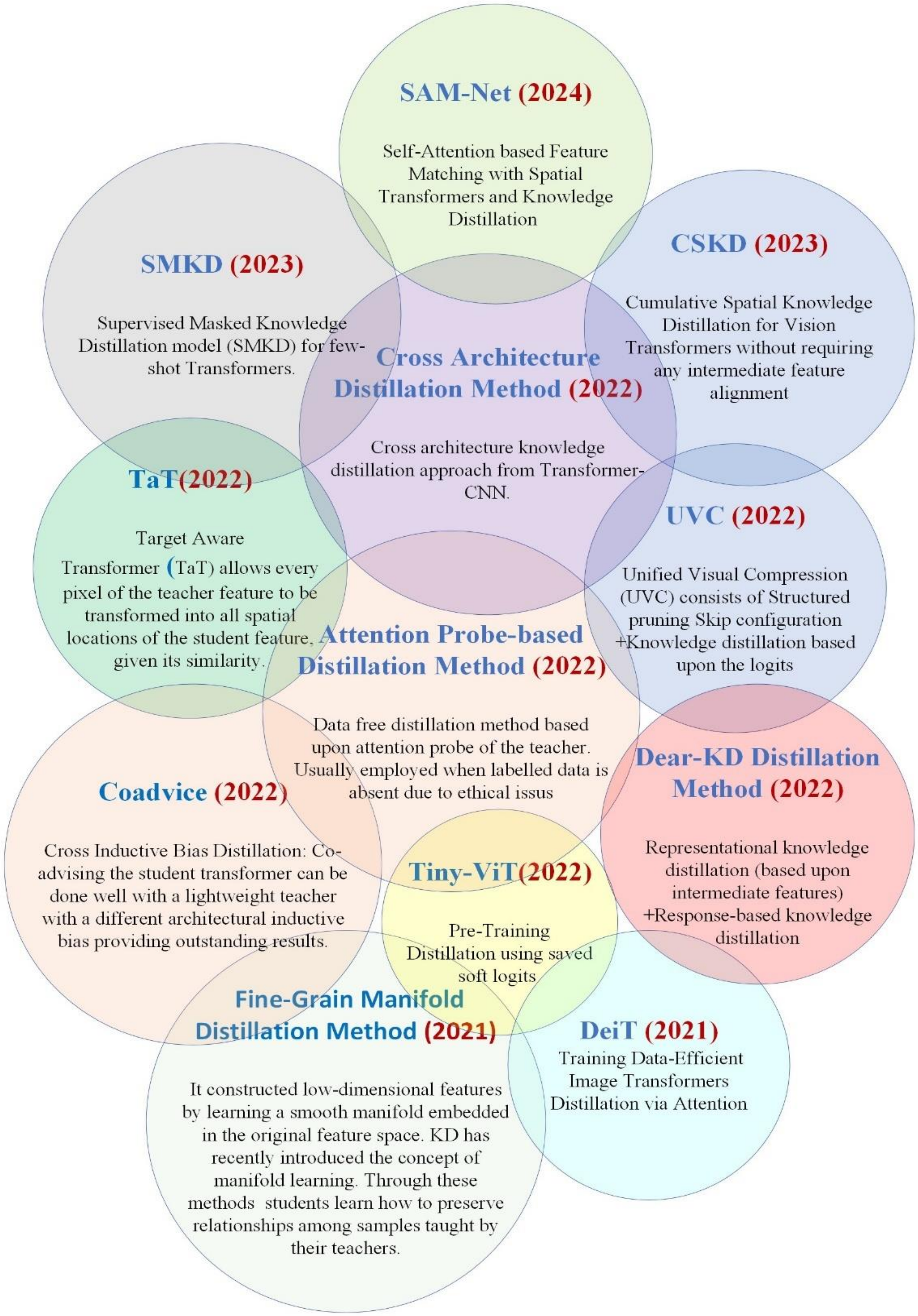

Figure 4: Taxonomy of KD approaches

### 4.1) Target aware Transformer

One of the novel solutions proposed by the authors [46] stated that in most previous studies, representational features are regressed from the teacher to the student as a one-to-one spatial match. However, because of the architectural differences, researchers need to pay more attention to this. A spatial location can have different semantic information depending on its location. As a result, the one-to-one distillation approach is undermined greatly. The authors in [46] accomplished this by proposing a novel one-to-all spatial matching knowledge distillation method through Target aware Transformer (TaT). A TaT allows every pixel of the teacher feature to be transformed into all spatial locations of the student feature, given its similarity. In their proposed method, the teacher's features were distilled into all student features using a parametric correlation, i.e., the distillation loss is a weighted summation of all student features. Using a transformer structure, they reconstructed each student feature component and aligned it with the target teacher feature. As a result, the proposed method was named TaT. One of the critical challenges faced by the proposed method was that it was likely to become intractable in the case of large feature maps since it calculated correlations between feature spatial locations. To overcome this, the authors extended the pipeline in a hierarchical two-step manner given as:

1) Rather than computing correlation across all spatial locations, they split the feature maps into several patches and performed one-to-all distillation within each patch.
2) They averaged the features within a patch into a single vector to distil the knowledge.
For feature alignment, TaT was used to transform the teacher, and student model features into similar feature spaces.

**Knowledge distillation approach:** Patch group distillation and Anchor point distillation with the Kull back divergence (KLD) loss function.

An illustration of the proposed methodology [46] is given in figure 3 and figure 4.

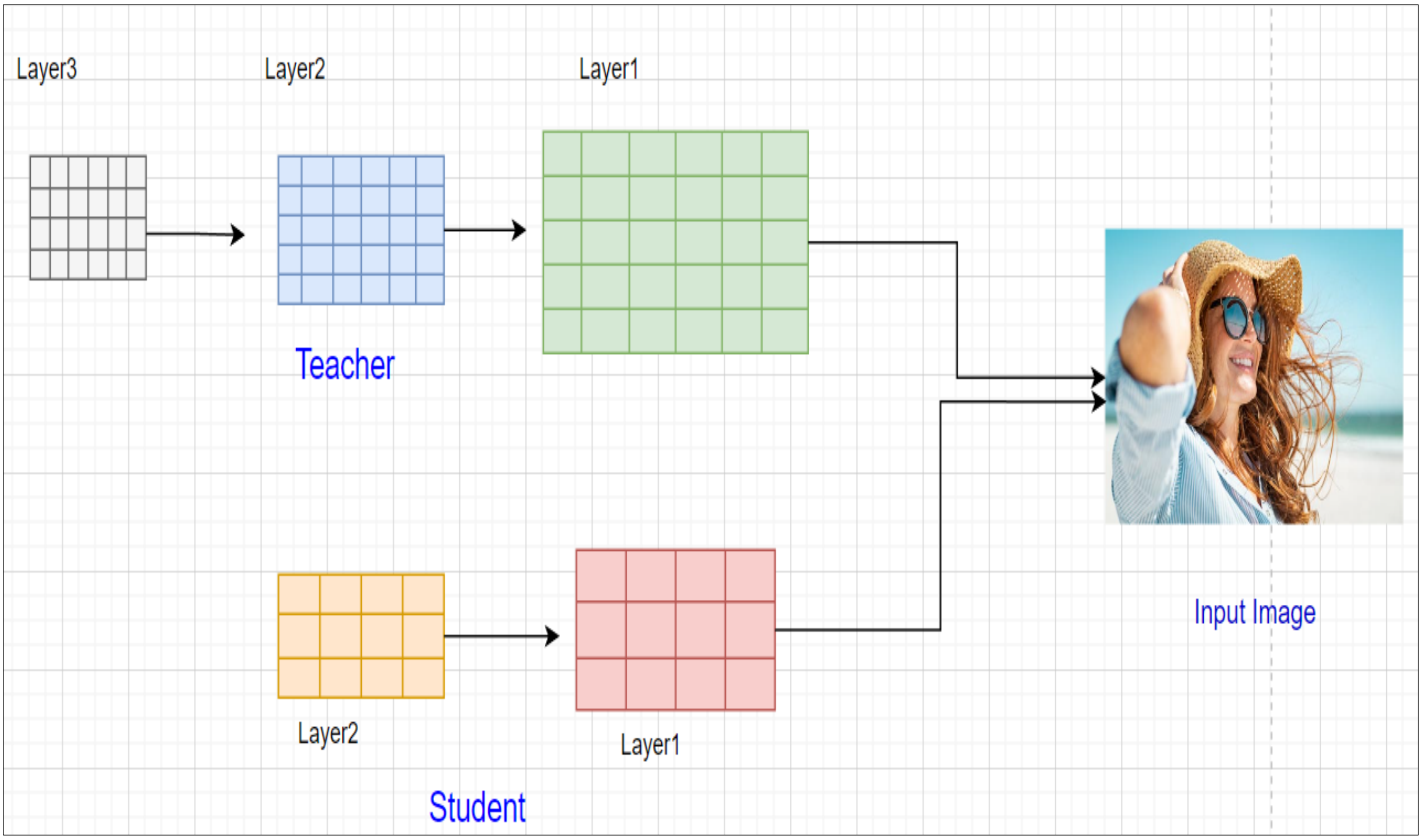

Figure 4 a) Compression of receptive fields

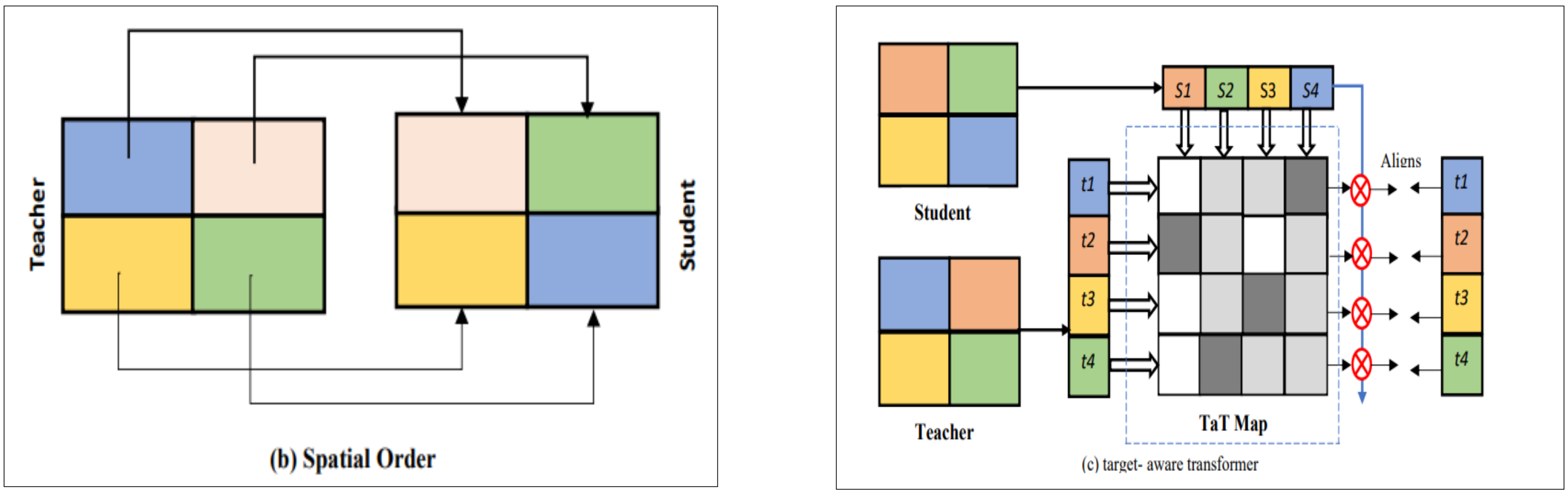

Figure 4. b) Illustration of semantic mismatch spatial order c) target aware Transformer.

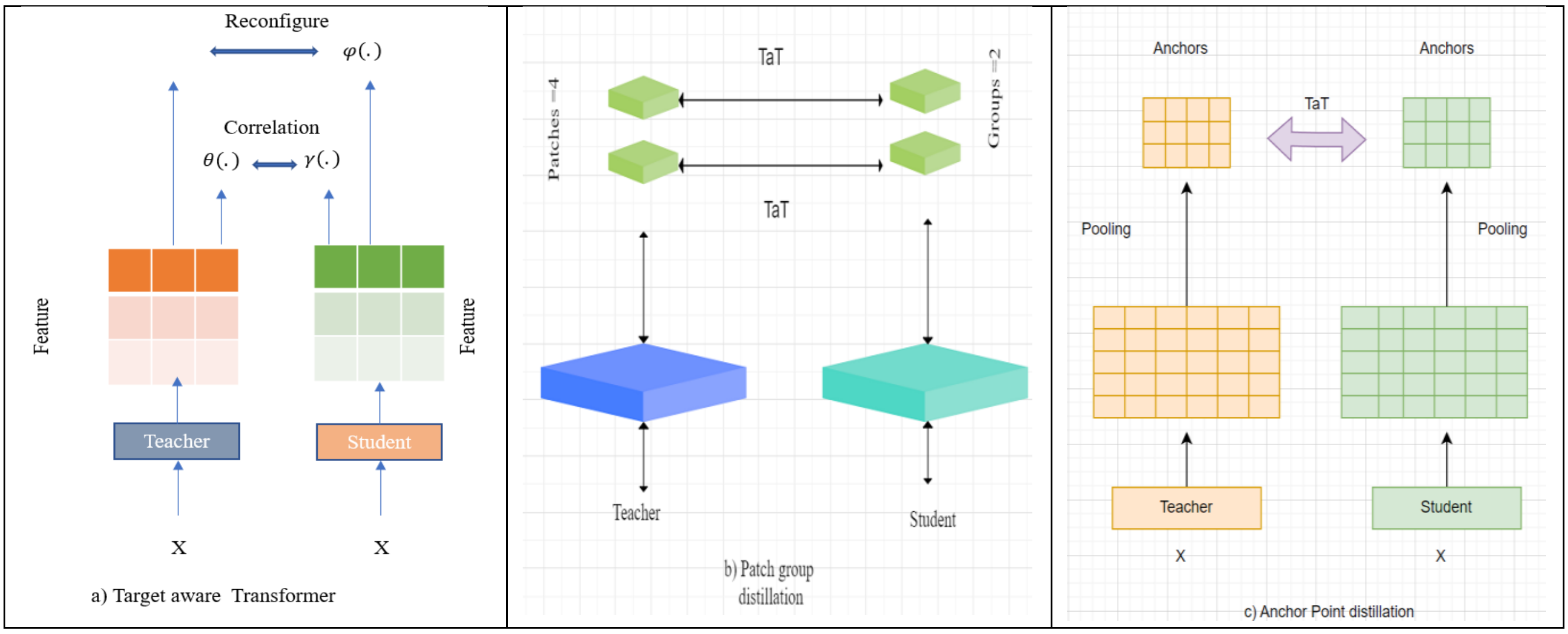

Figure 5. Illustration of Proposed Framework a) **The Target-aware Transformer** is based on the teacher and student features. The transformation map Corr. is computed and applied to the student feature to reconfigure itself to minimize the L2 loss with the teacher feature b). **The patch-group distillation** method requires slices of teacher and student features to be rearranged into groups for distillation. Adding patches to a group introduces the spatial correlation among the patches beyond the patches themselves). (c) **Anchor-point Distillation** using average pooling, Authors [46] extracted the anchor within a local area of the given feature map, forming a smaller feature map. Distillation was performed on the anchor-point features generated.

**Key observations**

- A student transformer can acquire more information by acquiring information from the intermediate layers in addition to logits.
- In [46], distillation was applied to the last layer of the backbone network, i,e, from the logits of the network.
- Some works have explored multi-layer distillation, so it would be interesting to see how effective it is when multiple layers are involved.

**4.2) Fine-Grain Manifold Distillation Method**

Another novel solution was given by the group of authors Zhiwei Hao [47], targeted learning via Fine-Grained Manifold Distillation. In manifold learning, dimensionality is reduced nonlinearly. It constructed low-dimensional features by learning a smooth manifold embedded in the original feature space. KD has recently introduced the concept of manifold learning. Through these methods [47], students learn how to preserve relationships among samples taught by their teachers. These primary attempts are coarse and can be further improved since patches instead of images are the basic input elements for ViTs. The proposed method decouples a manifold relation map into three parts using the orthogonal decomposition of matrices. There are three parts to the relation map, an intra-image relationship map, an inter-image relationship map, and a randomly sampled relationship map. For example, in the paper [47], a manifold relation map is calculated across each group of patches containing the same colour, as shown in figure 5.

**Knowledge distillation approach:** Manifold distillation loss (MD Loss) is used, which is a combination of intra-image patch level distillation loss +inter image patch level distillation losss+random sampled patch level manifold distillation loss together with original Knowledge distillation loss function. Manifold distillation loss is computed to map the teacher and student model features. Figure 5 illustrates the image level and patch level manifold as follows:

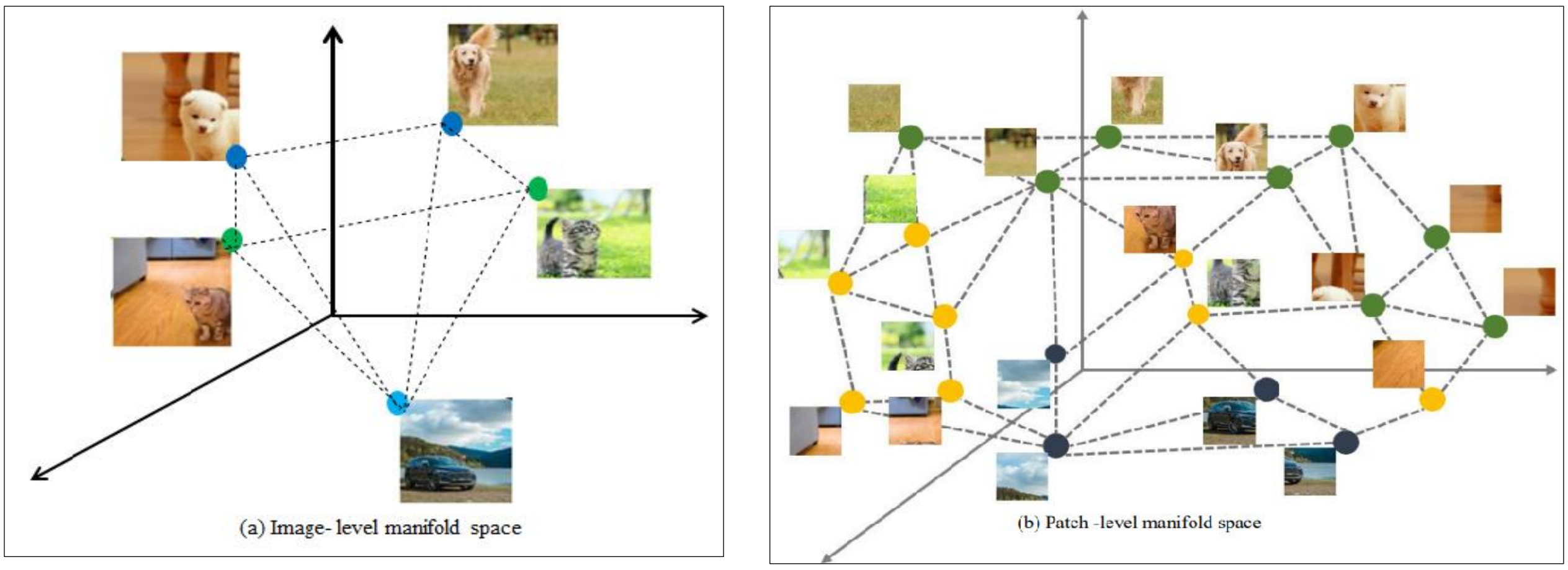

Figure 6. Comparison between (a) image-level and (b) patch-level manifold spaces. Knowledge transfer is facilitated by patch-level manifold spaces containing fine-grained information.

The general architecture of the proposed method for manifold distillation is given in figure 6:

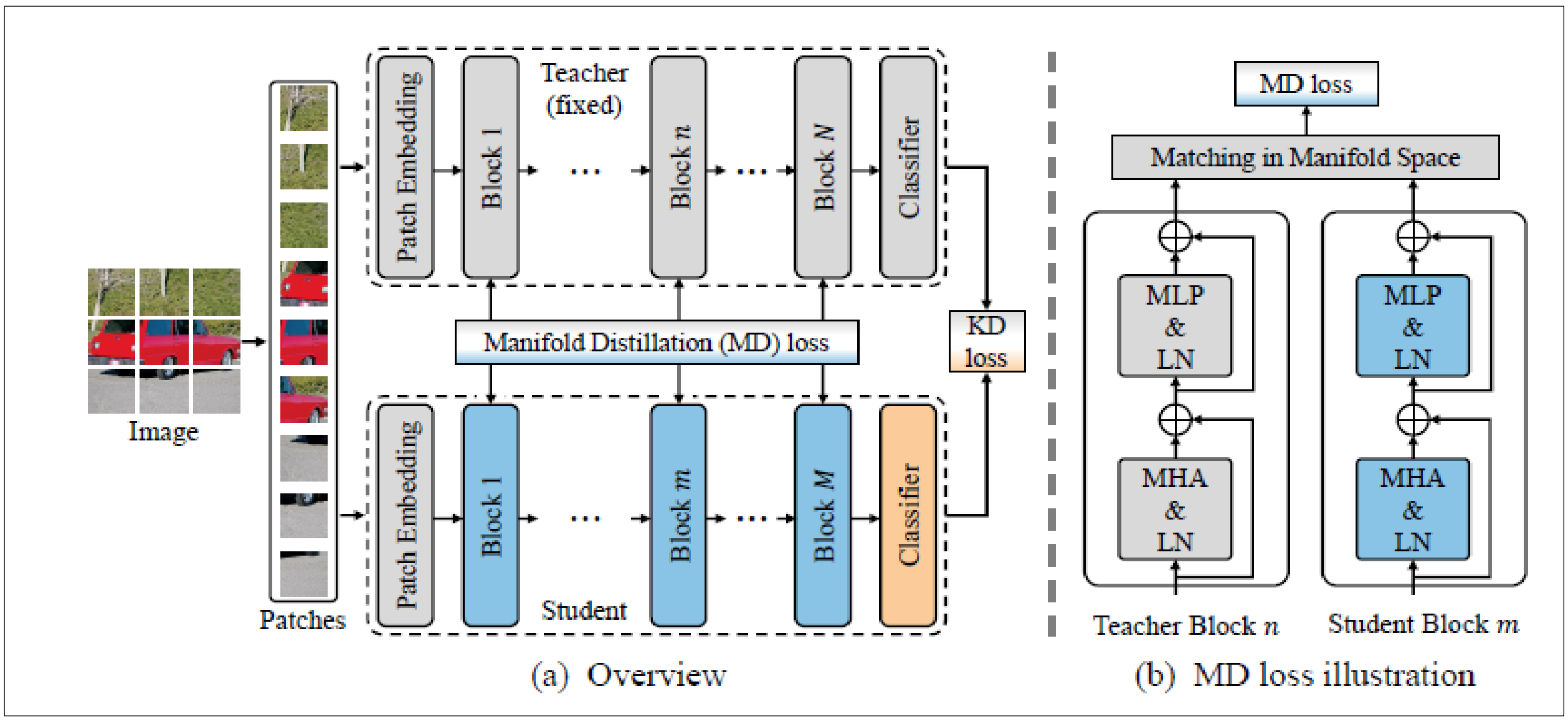

Figure 7. This method involves the fine-grained distillation of manifolds. a). As a method of transferring knowledge from one teacher to another, the manifold distillation loss is used in conjunction with the original KD loss) The manifold distillation loss is computed by matching feature relationships between each pair of selected teacher-student layers in manifold space.

**Key observations:**

- A manifold space mapping student and teacher features are used, and the mapping loss is decoupled into three terms to reduce computational complexity.
- Decoupling significantly reduces computation and memory space, but the computing and storing overhead is too high if the patch size is too small.

- The patch number N in the Swin transform is 3136 for an input size of 224×224. Using such a large patch number significantly increases the computational complexity and memory space requirement of the intra-image manifold loss Lintra.

### 4.3) Cross Inductive Bias Distillation (Coadvice)

Sucheng et al. [48] proposed another novel solution named Cross Inductive Bias Distillation (Coadvice). The authors in the paper [48] stated that teacher accuracy is not the dominant factor in student accuracy, but inductive teacher bias is. Co-advising the student transformer can be done well with a lightweight teacher with a different architectural inductive bias providing outstanding results. As a result, teachers with different inductive biases possess diverse knowledge despite being trained on the same dataset, and models with different inductive biases tend to focus on diverse patterns. During distillation, the student gained a more accurate and comprehensive understanding of the data and compounds because of distilled diverse knowledge by the multi-teacher network. A token inductive bias alignment was also proposed to align the inductive bias of the token with its target teacher model. Proposed vision transformers (CiT) in [48] outperformed all existing ViTs using only lightweight teachers with cross-inductive bias distillation methods.

**Key observation points**

- According to the study, teachers' intrinsic inductive bias matters more than their teacher accuracy.
- CNN and INNs have inductive biases, which lead to complementary patterns, whereas vision transformers with fewer inductive biases can inherit information from both architectures.
- Students who are given multiple teachers with different inductive biases are more likely to learn a variety of knowledge when those teachers have different inductive biases.
- Knowledge distillation makes student transformers perform similarly to various inductive bias teachers compared to introducing inductive bias into the Transformer.
- A cross-inductive bias vision transformer (CiT) was developed in this study that outperformed all existing vision transformers of the same architecture. The super lightweight teachers used in this study had only 20% and 50% of the DeiT-Ti and DeiT-S parameters, respectively.

**Knowledge distillation approach:** Convolution +involution inductive bias knowledge distillation approach.
**Loss function:** Kull back divergence loss function + Cross entropy loss function. The overall architecture of the proposed method is best illustrated by the Figure given:

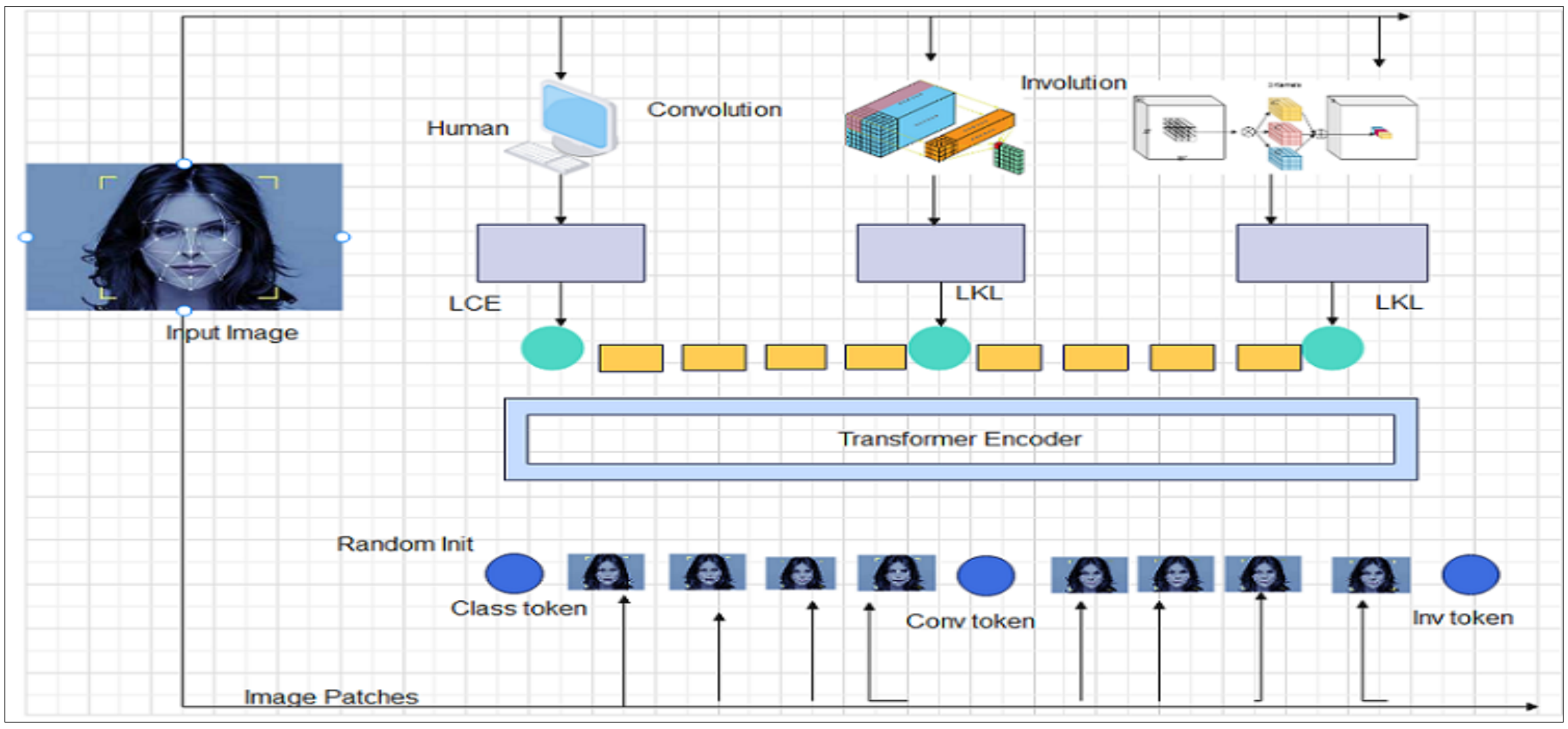

Figure 8. Cross Inductive bias knowledge distillation framework. a) Taking an image as input, the human, convolution, and involution models will give three similar but slightly different distributions to describe the class of images. Compared to the ViT model, the CiT model has two additional tokens (i.e., convo and into tokens) from the convolution and involution teacher.

### 4.4) Pre-Training Distillation using saved soft logits (Tiny-ViT)

Kan Wu et al. [49] proposed a different solution known as Tiny-ViT compared to the previously mentioned methods focused most of the time on finetuning distillation. They observed that small pre-trained models could benefit from knowledge transferred from large to small pre-trained models using massive pretraining data. In particular, the authors employed distillation during pretraining to transfer knowledge. It is possible to save memory and computation overhead by sparsifying and storing logits in advance for large teacher models. A large pre-trained model with computation and parameter constraints is automatically scaled down to produce tiny student transformers. Extensive experiments have demonstrated the effectiveness of Tiny-ViT. It achieved top-1 Accuracy of 84.8% using only 21M parameters, comparable to Swin-B without using as many parameters as Swin-B. Tiny-ViT also achieved 86.5% accuracy with increased image resolutions, slightly exceeding Swin-L's Accuracy with only 11% parameters.
The proposed work in the paper [49] focused on pretraining distillation, making small models more capable of learning from larger models and transferring those skills to downstream tasks.
The critical challenge of pretraining distillation methods is that huge computing resources are wasted in passing training data through large teacher models instead of training a small target group. The distillation method is inefficient and expensive. A large-scale teacher may also consume a large amount of GPU memory, slowing training speeds for target students.
The authors proposed a framework [49] that allows fast pretraining distillation to solve this problem. Teachers' predictions and data augmentations are stored in advance by the authors. As a result, Training simplifies the distillation procedure by reusing stored information, so the large teacher model does not require as much memory and computation. Authors were required only to recover the augmentation information and teacher prediction from stored files and optimize the cross entropy objective function for student model distillation.

**Key observations**

- Because authors only used the soft labels generated by teacher models for Training, the proposed framework is label-free, i.e., without the need for ground-truth labels.
- Therefore, it can utilize a wide range of off-the-shelf web data without labels for large-scale pretraining.
- In practice, such a strategy is feasible since soft labels carry enough discriminative information for classification, including category relations, and are accurate enough to be used.
- However, distillation with ground truth would result in slight performance degradation.

**Knowledge distillation approach:** Distillation from stored sparse soft logits and pre-stored data augmentation information.
**KD loss function: Cross entropy loss function:** Generic architecture of the proposed method is given in Figure 8 as follows:

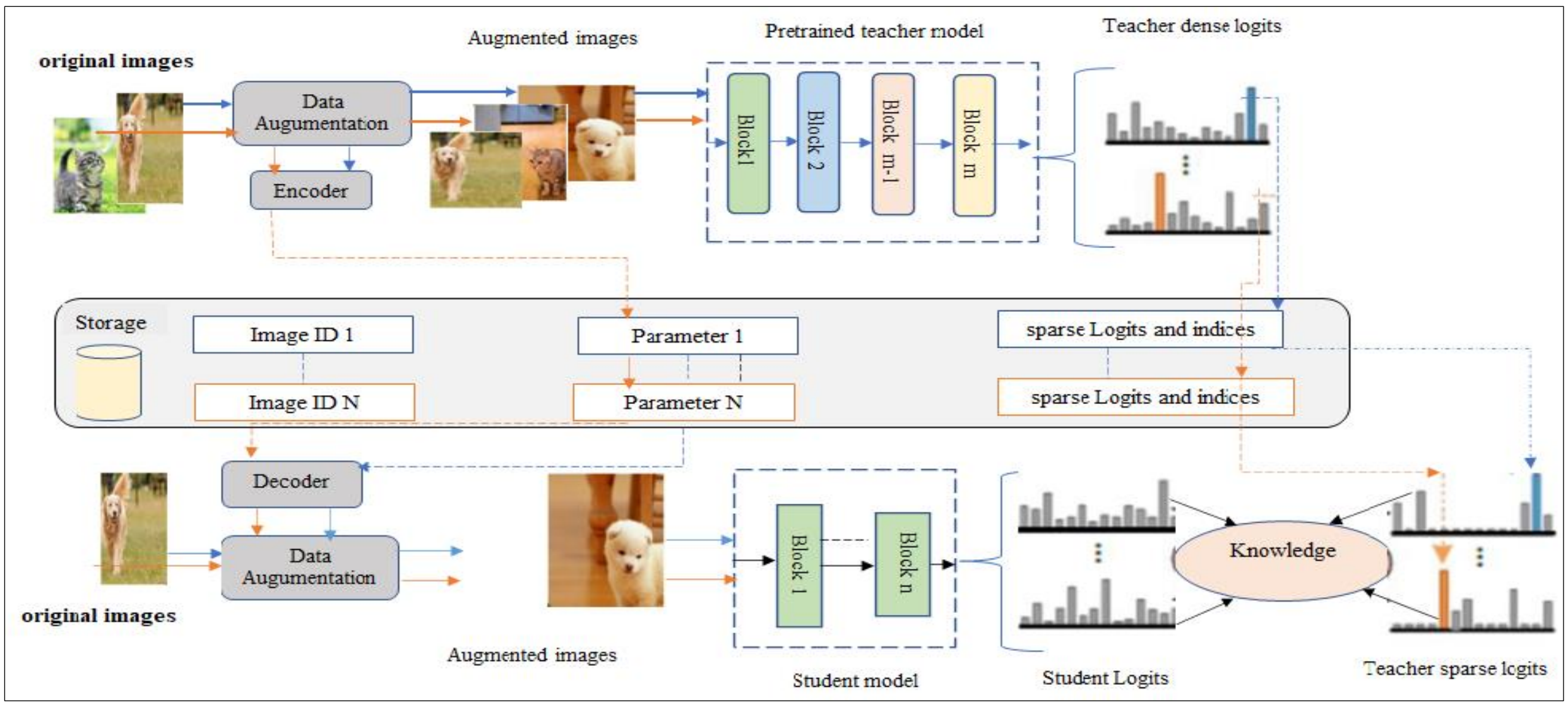

Figure 9. Pretraining distillation framework for fast Training **Top:** branches for saving teacher logits are shown at the top. Sparsified teacher logits and encoded data augmentation are saved. **Middle**: A disk is used to store information in the middle.

**Bottom:** The branch for training students is at the bottom. The decoder distilled the teacher logits into student outputs by reconstructing the data augmentation. It should be noted that each branch runs independently and asynchronously.

### 4.5) Attention Probe-based Distillation Method

The author Jiahao Wang [50] suggested a novel solution known as the Attention-Probe based upon the consideration when the true distribution of data is not available due to ethical and legacy issues. Due to higher computational requirements, ViTs are incompatible with resource-constrained devices. Compressing them using the original training data might be possible, but privacy and transmission issues may prevent that. Many unlabelled data in the wild offered an alternative paradigm for compressing convolutional neural networks (CNNs). However, a similar paradigm for ViTs remains an open question due to the significant differences between CNNs and ViTs in model structure and computation mechanism. This paper aimed to compress ViTs using two stages of unlabelled data collected in the wild. The first step was to design a tool to select valuable data from the wild, an Attention Probe. The second step used the selected data to train a lightweight student transformer using a probe knowledge distillation algorithm, which maximized the similarity between the heavy teacher and lightweight student models in terms of both outputs and intermediate features. The attention probe is illustrated by the Figure given:

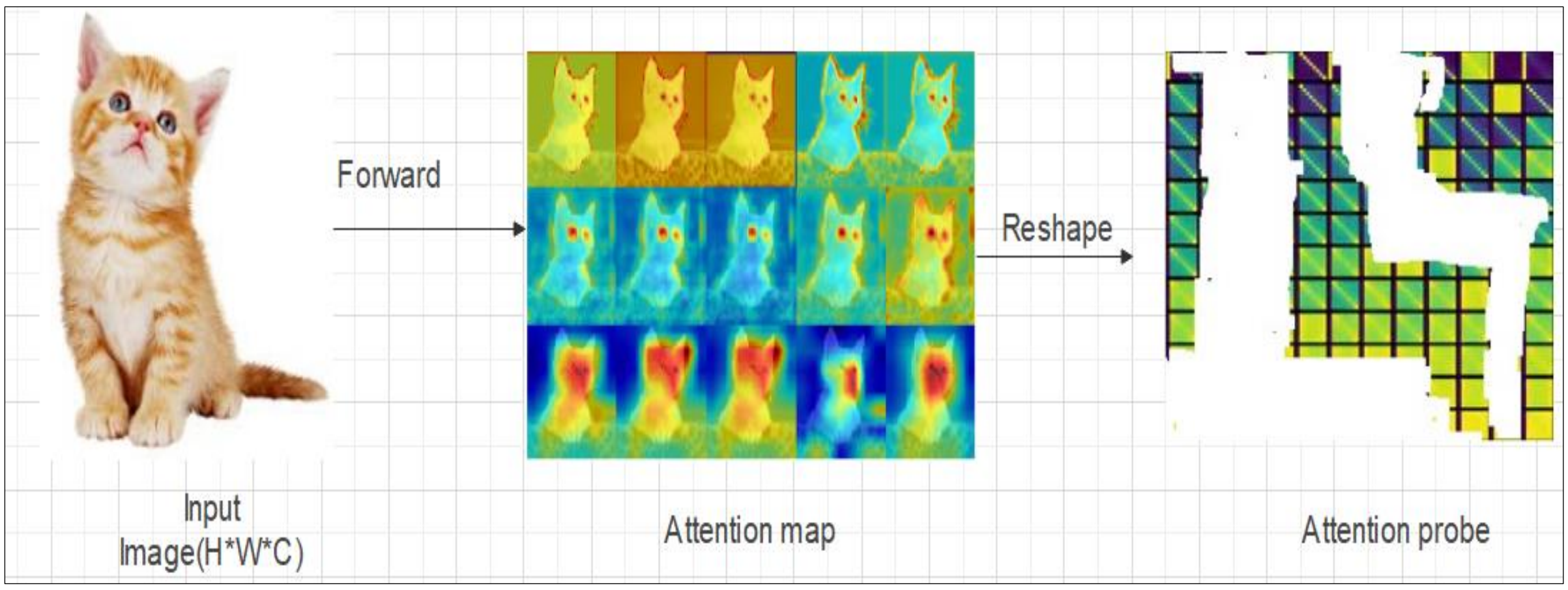

Figure 10. Attention probe calculation paradigm

Experiments on several benchmarks showed that the student transformer obtained by the proposed method was comparable to the baseline based on the original training data. The attention probe is defined as:

**Attention probe:** Our attention probe is a special section of the attention map based on random input $x^U$ from the wild data.

$$P\left(X^U\right)=\Psi\left(A\left(X^U\right)[0:N+1]\right) \tag{12}$$

Where $P \in R^{H*W}$ , $N = H \times W$ , and $\Psi$ the reshape operation converts $R(1\times N)$ to $R(H \times W)$. As a result of the self-attention mechanism, the forward propagation of the image first obtains the attention map, and then the last $N$ elements of the first row are extracted and reshaped. Using the attention probe, they determined how much attention the model's [class] token pays to the other $N$ patches. This way, it collected data from a large volume of unlabelled wild data.

From this study [50], it is straightforward to understand how the attention value works. The recognition capacity for data near the original data distribution is improved with a teacher transformer trained on the original data. To perform the recognition task, the [class] token of the teacher determined which area of the input image should be emphasized. Therefore, the variance of values in an attention probe is often large and much lower than the variance in a uniform matrix SN. Conversely, the pretrained teacher's variance is lower and closer to SN when the sampled data come from the original distribution. Due to this limitation, many out-of-distribution jamming data cannot be collected.
**Knowledge distillation approach:** Probe distillation + Knowledge distillation.
**KD loss function:** Probe distillation function for distilling intermediate features and cross-entropy.
The generic architecture of the proposed methodology is given in Figure 10. as:

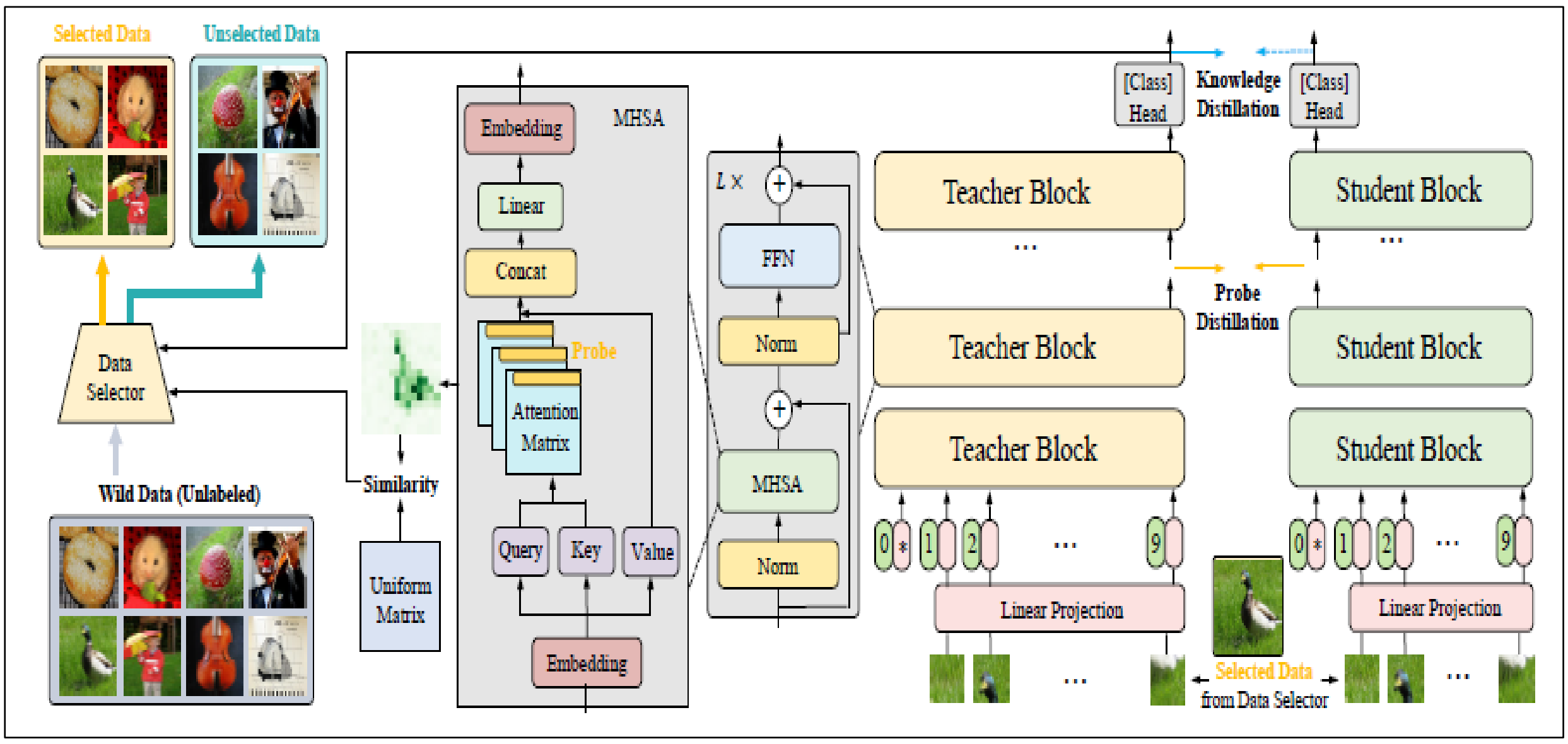

Figure 11. An illustration of how student vision transformers are learned in the wild. Using the output from the teacher network and attention probe, the data selector picked out useful data from the wild data and used it for the next distillation process. Probe distillation is used to extract information from the intermediate features of the heavy Transformer that has been pre-trained.

**Key observations**

- Motivation in the data-free scenario is the lack of supervised information since the collected data is not labelled.
- This inevitably results in information loss and accuracy degradation due to the lack of cross entropy between student output and ground-truth labels.
- To compensate for the loss of supervised information, authors considered the intermediate information gained from pre-trained teacher transformers.
- In addition to outputs, the intermediate layers contain more embedding, enabling the student transformer to acquire more information.

### 4.6) Data-Efficient Image Transformers Distillation via Attention

Touvron et al. [51] presented a novel technique named Training Data-Efficient Image Transformers Distillation via Attention. The authors proposed a teacher-student strategy based on distillation tokens specific to transformers. A distillation token ensures that the student learns from the teacher through attention, generally from a ConvoNet teacher. On ImageNet, the learned transformers perform competitively (85.2% top-1 accuracy) with state-of-the-art and similarly on other tasks. The distillation procedure consists of a distillation token that plays the same role as the class token, except that it aims to reproduce the label estimated by the teacher. As a result of attention, both tokens interact in the Transformer through this mechanism. By a considerable margin, this transformer-specific strategy is superior to vanilla distillation. In this way, the authors [51] showed that neural network models lacking convolutional layers could achieve comparable results against state-of-the-art on benchmark datasets such as ImageNet with no external data. They [51] also observed that in comparison to ResNet-50 and ResNet-18, their two novel models, DeiT-S and DeiT-Ti, achieved better performance with fewer computational requirements.
**Knowledge Distillation Approach:** Distillation through attention using a distillation token pre-learned from the ConvoNet teacher. True label and teacher prediction was used to finetune the model at a higher resolution.

**KD loss function:** Kull back divergence distillation loss function + Cross entropy distillation loss function. The distillation procedure is illustrated in figure 11.

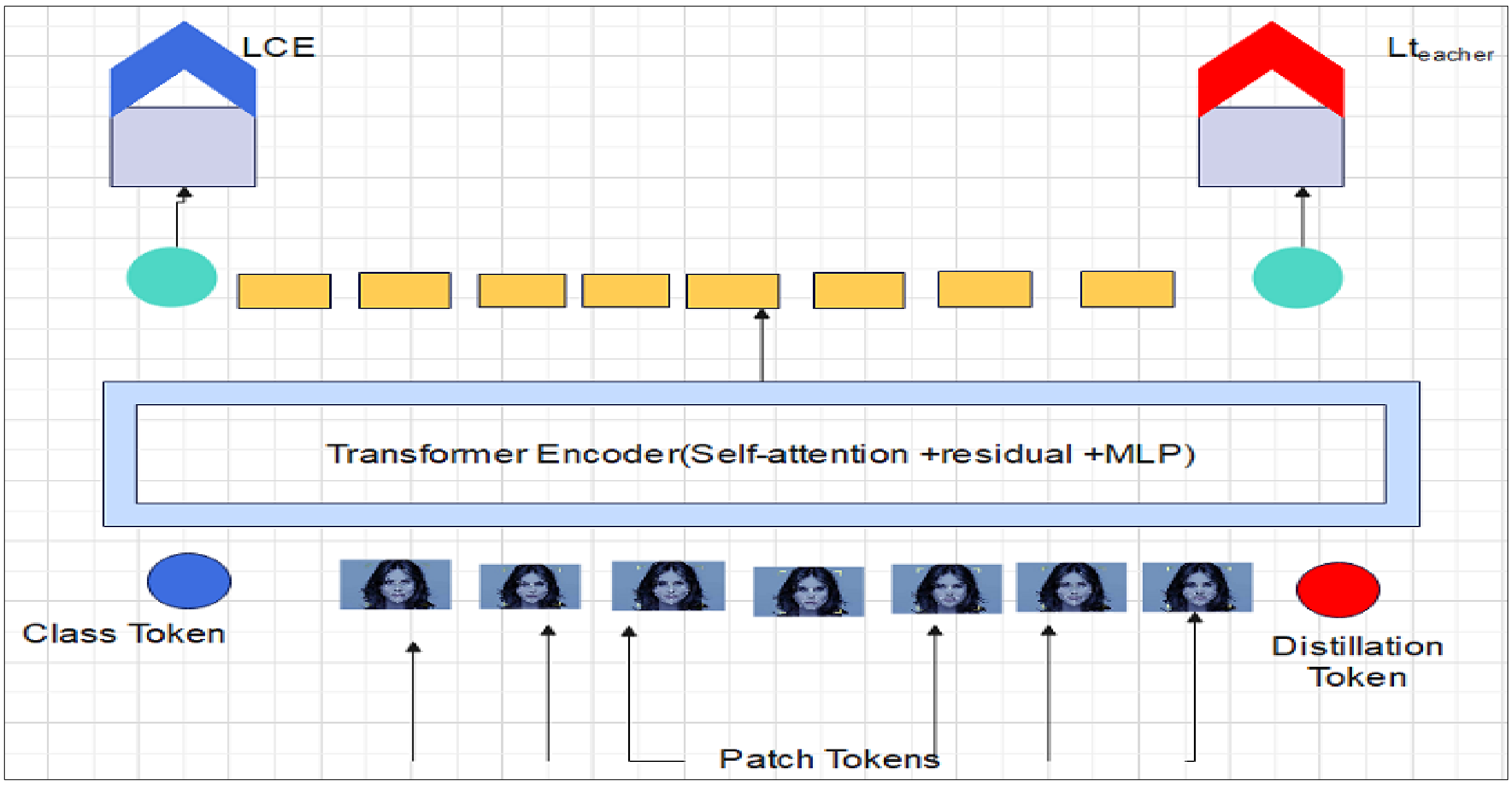

Figure 12. The distillation procedure depends upon the distillation Token. The distillation token interacts with the class and patch tokens as part of the self-attention layer. On the network's output, distillation tokens are employed the same way as class tokens, except that they aim to replicate the (hard) labels predicted by the network. Backpropagation allows the transformers to learn the classes and distillation tokens.

### 4.7) Unified Visual Transformer Compression (UVC)

The authors, Shixinng Yu et al., in this article [52], presented a unified framework called Unified Visual Transformer Compression (UVC) that seamlessly integrated pruning, layer skipping, and knowledge distillation. Under a distillation loss, the Authors developed an end-to-end optimization framework aiming to jointly learn model weights, layer-wise pruning ratios, and skip configurations. Primitive-dual algorithms are then used to solve the optimization problem. The authors in this paper [52] used the ImageNet dataset to test several ViT variants, including a DeiT and a T2T-ViT backbone, and their method consistently outperformed previous competitors. DeiT-Tiny's FLOPs were reduced to 50% of the original FLOPs without compromising Accuracy. Using this proposed unified framework, they aimed to prune each layer's head number and dimension simultaneously associated with the layer level skipping. According to our knowledge from this study, they do not extend the scope of the reduction to other dimensions, such as the number of input patches or the size of tokens. These parts can, however, also be easily bundled together using this unified framework.
**Knowledge Distillation Approach**: Unified Visual Compression (UVC) consists of Structured pruning Skip configuration +Knowledge distillation based upon the logits.
**KD loss function:** l2 norm distillation loss over KL divergence distillation loss function.
**Key observations:**

- Implementing only skip connection manipulation will cause high levels of instability. Due to the large architecture changes (adding or removing one whole block), the objective value fluctuates heavily during optimization.
- Furthermore, using only skip manipulation, the Accuracy is remarkably reduced, e.g., 4% on DeiT-Tiny.
- Pruning within a block-only method performs much better than skip manipulation due to its finer-grained operation.
- However, the method is still behind the proposed joint UVC method since the latter also removes block-level redundancy a priori, which has recently been widespread in finetuned transformers.

### 4.8) Dear-KD Distillation Method

The authors Xiang Chen et al., in this study [53], considered the situation where true data distribution of data is not available, known as Dear-KD. The powerful modelling capacity of transformers with self-attention makes them ideal for computer vision applications. Despite this, the excellent performance of transformers relies heavily on enormous training images. It is, therefore, urgent to develop a data-efficient transformer solution. In this study [53], the Authors suggested an early knowledge distillation method.

The proposed framework, known as the Dear-KD framework, is designed to enhance the data efficiency required of transformers. Dear KD, as a two-stage framework, distilled the biases in stage I from the early intermediate layers of a CNN and gave the Transformer full rein by Training without distillation in second stage II. To further reduce the performance gap versus the full-data counterpart, the Authors [53] proposed a boundary-preserving intra-divergence loss based on Deep Inversion. They demonstrated the superiority of Dear-KD over its baselines and state-of-the-art methods on ImageNet, partial ImageNet, data-free settings, and other downstream tasks. Convolution in the early layers of the proposed network has been shown to enhance the performance significantly. Since local patterns (like texture) were captured well in the earlier layers of the model, it was, therefore, necessary to provide explicit feedback to the early transformer layers regarding inductive biases to improve data efficiency. However, when the Transformer reached its later phases, this guidance limited its ability to express itself fully. Therefore, the Transformer was given full rein in stage II to express its modelling capacity fully. They achieved state-of-the-art image classification performance with comparable or fewer computations using the full ImageNet. It was impressive to observe how Dear-KD outperformed the baseline transformer trained with all ImageNet data with only 50% of the data. Finally, the DeiT-Ti-based DearKD achieved 71.2% on ImageNet, which is only 1.0% lower than the full-ImageNet DearKD.

**Knowledge Distillation Approach**: Representational knowledge distillation (based upon intermediate features) +Response-based knowledge distillation. The Dear-KD is also explored in the extreme setting (data-free) where there is no access to real images called by the name DF-DearKD. Compared with DearKD, DF-DearKD consists of an extra mage generation component based upon Deep Inversion [53] and Boundary preserving intra-divergence loss [53].

**KD loss function:** MSE distillation loss for hidden features+, Cross entropy loss for hard label distillation, and intra-divergence distillation loss function for the data-free environment.

**Key Observations**

- Due to the limited information stored in feature statistics, DF-DearKD cannot handle human-related classes despite being able to generate high-quality images.
- Furthermore, generating many samples requires much time and computation as the study doesn’t use real images.
- Training with generated samples still needs to be equivalent in performance to Training with real images.

The pipeline for the DF-DearKD is given in Figure 14 as follows:

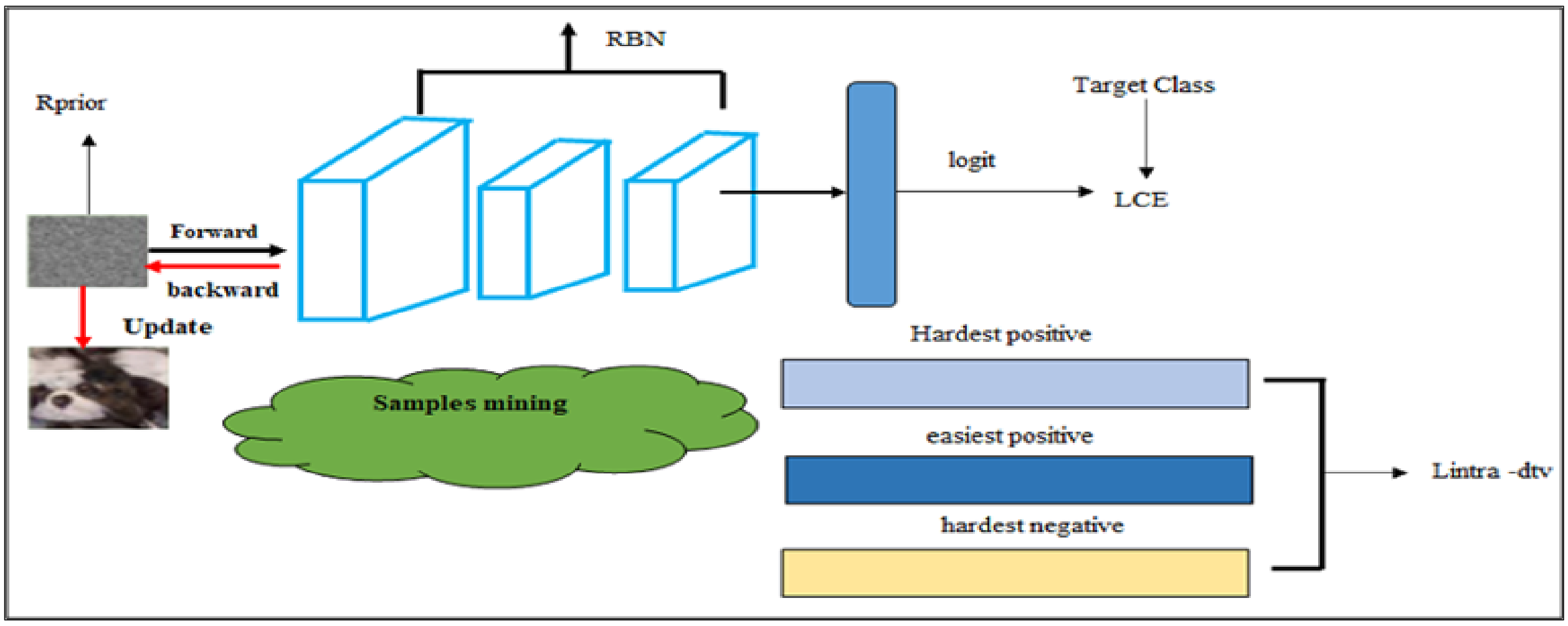

Figure 14. Pipeline for DF-Dear-KD method

### 4.9) Cross Architecture Distillation Method

Most of the solutions discussed above considered homologous architectures for distillation purposes, which is only sometimes an optimal solution. Yufan Liu et al. [54] proposed cross-architecture knowledge distillation methods to combat heterogeneous architectures' gaps. Due to Transformer's superior performance and ability to understand global relations, Transformer has attracted great attention. A CNN can be used to distil complementary knowledge from Transformer for higher performance. Most knowledge distillation methods use homologous-architecture distillation, such as distilling knowledge from CNN to CNN. They may not be appropriate in cross-architecture scenarios, such as from Transformer to CNN.

The purpose of the study [54] was to propose a novel method for distilling knowledge between cross-architectures. A significant increase in knowledge transferability existed when Transformer was used to transfer knowledge to a student CNN network. In a transformer teacher model, students learned local spatial features (from a CNN model) and complementary global features (from a CNN model). Two projectors were designed, one for partial cross-attention (PCA) and one for group-wise linear (GL). These two projectors aligned students' intermediate features into two different feature spaces, making it easier to distil knowledge rather than directly emulating a teacher's output. Using the PCA projector, the student feature was mapped to the teacher's Transformer attention space. A Transformer teacher used this projector to explain how global relationships work to students. To create a Transformer feature space, each student's feature is mapped to the Transformer feature space using GL projectors. Therefore, it helped to alleviate the differences between the ways the teacher and the student form feature.

Furthermore, the authors presented a robust multi-view training scheme to enhance the stability of the framework and robustness. Experimental results have shown that the proposed method performed better on large and small datasets than 14 state-of-the-art methods.

The paper ss also proposed a robust adversarial cross-view training scheme to mitigate the instability caused by cross-architecture diversity. Samples representing multiple views are necessary for disrupting student networks. They constructed an adversarial discriminator that can be used to distinguish between teacher and disturbing student features. In comparison, teaching the student to confuse the discriminator. Students are capable of becoming more stable and robust after experiencing convergence.

**Knowledge Distillation Approach**: Cross architecture knowledge distillation approach from Transformer-CNN.

**Feature alignment:** Partially cross attention (PCA) projector and Groupwise Linear (GL) projector Promotes teacher features' transferability by aligning the student feature space.

**KD loss function:** GL loss function + PCA loss function + for multi-view generative adversarial training MVG loss function is calculated. Finally, the vanilla KD loss function is used for distillation.

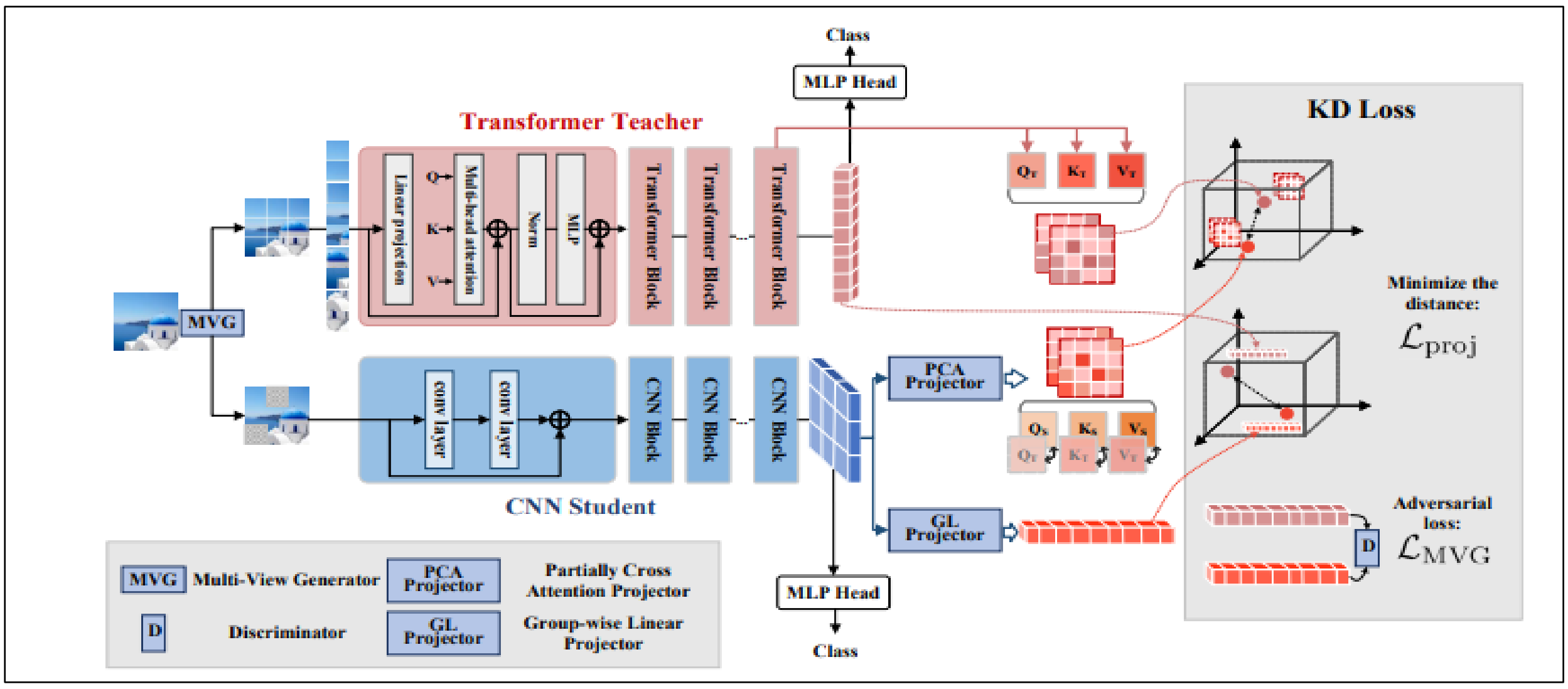

Figure 15. Cross-architecture knowledge distillation framework

**Key observations**

- During the KD procedure, the two projectors gained better performance on ImageNet.
- As a result, PCA and GL projectors significantly improved the quality of CNN features, although they were removed during inference.
- There was a large increase in cosine similarity, even higher than in homologous architectures.
- Consequently, carefully designed KD methods can result in a higher knowledge transfer between the Transformer and CNN.

### 4.10) Supervised Masked Knowledge Distillation model (SMKD) for few-shot Transformers

In this [56-57] paper, the authors presented a pioneering Supervised Masked Knowledge Distillation (SMKD) method tailored for few-shot transformers that leverages recent advances in self-supervised knowledge distillation and masked image modeling (MIM). By integrating label information, this method enhances self-distillation frameworks, enabling intra-class knowledge distillation on both a class level and patch token level. The task of reconstructing masked patch tokens across intra-class images is one of the novel aspects of their approach. Extensive testing of their straightforward but powerful few-shot classification method proved to significantly surpass previous models, establishing a new benchmark. Further evidence using ablation studies in this [56-57] paper highlighting the powerful contributions of each component within our model can be found in comprehensive ablation studies. The overall framework of the distillation method is given in figure 16.

**Knowledge Distillation Approach**: Supervised masked KD approach combining cls token distillation + masked patch token distillation, with few shots learning evaluation.

**KD loss function:** Contrastive loss function (Cls Loss+ Patch token loss+ MIM loss).

**Key Observations**

- Incorporation of label information into self-distillation
- Allows intra-class KD for cls and patch tokens.
- An introduction to the challenging task of detecting patch tokens across intra-class images using masked patch tokens.

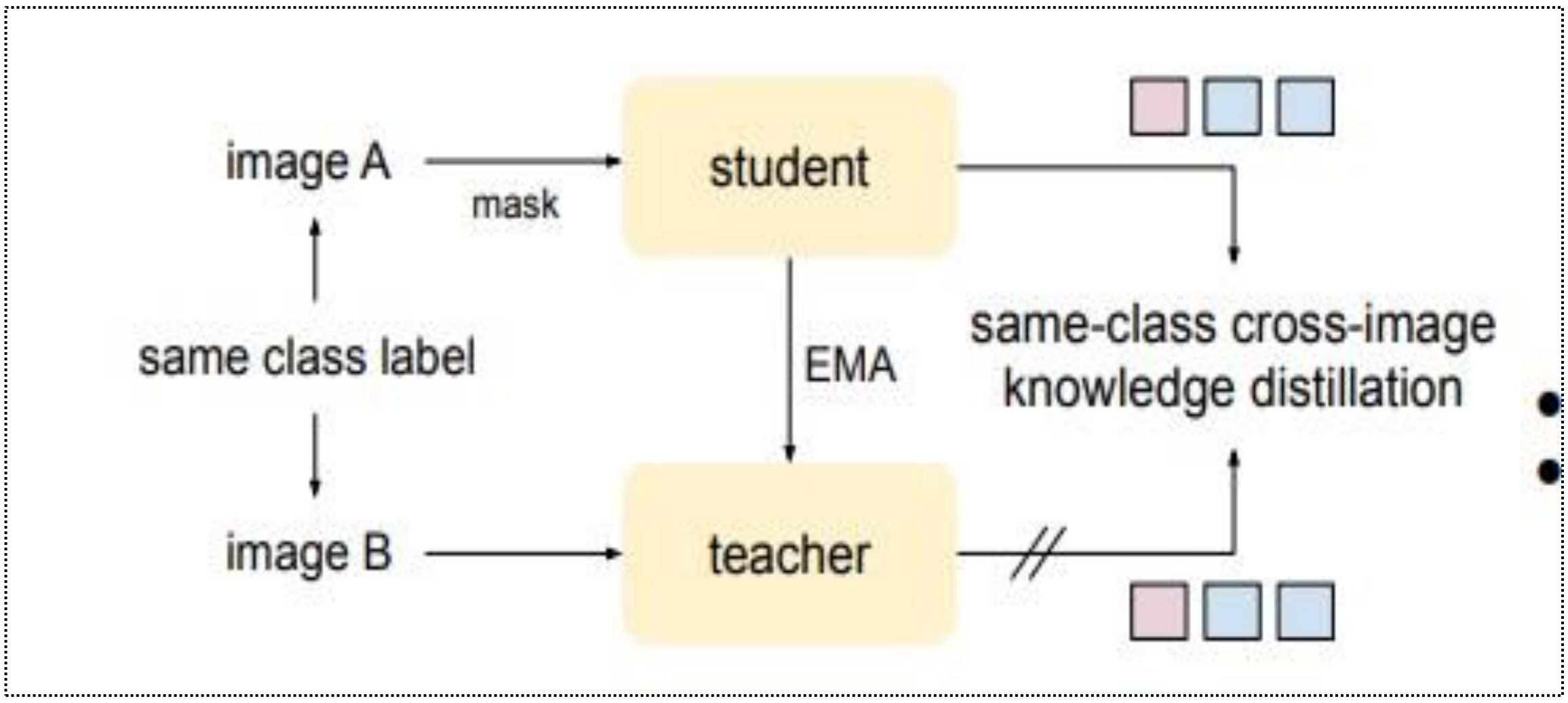

Figure 16. Through the integration of masked knowledge distillation with a supervised context, the model is able to address the overfitting problem in few-shot Transformers. Through this, it aligns [cls] with their respective [patch] tokens within the same class across images.

### 4.11) Cumulative Spatial Knowledge Distillation for Vision Transformers

The authors of this study [58-59] identify two key challenges hindering the effective distillation of knowledge from CNNs to Vision Transformers (ViTs): the difficulty of aligning spatial features between CNNs and ViTs, and the constrained capability of ViTs at later stages of learning. In order to overcome these issues, the authors [58-59] introduce the concept of Cumulative Spatial Knowledge Distillation (CSKD). Using this approach, ViTs are able to produce patch tokens directly from CNNs, without requiring any intermediate feature alignment. Moreover, the authors developed a Cumulative Knowledge Fusion (CKF) module that exploits local inductive biases from CNNs during early training phases and maximizes ViT's potential towards the end of training. The authors [58-59] validated the effectiveness and superiority of CSKD across ImageNet-1k and several downstream tasks through comprehensive experiments and analyses.

**Knowledge Distillation Approach:** With Cumulative Spatial Knowledge Distillation (CSKD), spatial knowledge is directly transferred from CNNs to all ViT patch tokens without the need for intermediate feature representations.

**KD loss function:** Kull back divergence distillation loss function + Cross entropy distillation loss function.

**Key Observations**

- In contrast to traditional feature representations, CSKD extracts spatial knowledge directly from CNNs to ViT patch tokens.
- As a result of Cumulative Knowledge Fusion, ViTs are enhanced by utilizing CNN-derived local-inductive-bias knowledge and maximising ViT capabilities over time.
- ImageNet-1k and various downstream tasks validate the superior performance of CSKD.

The overall framework of the proposed methodology can be illustrated in figure 17.

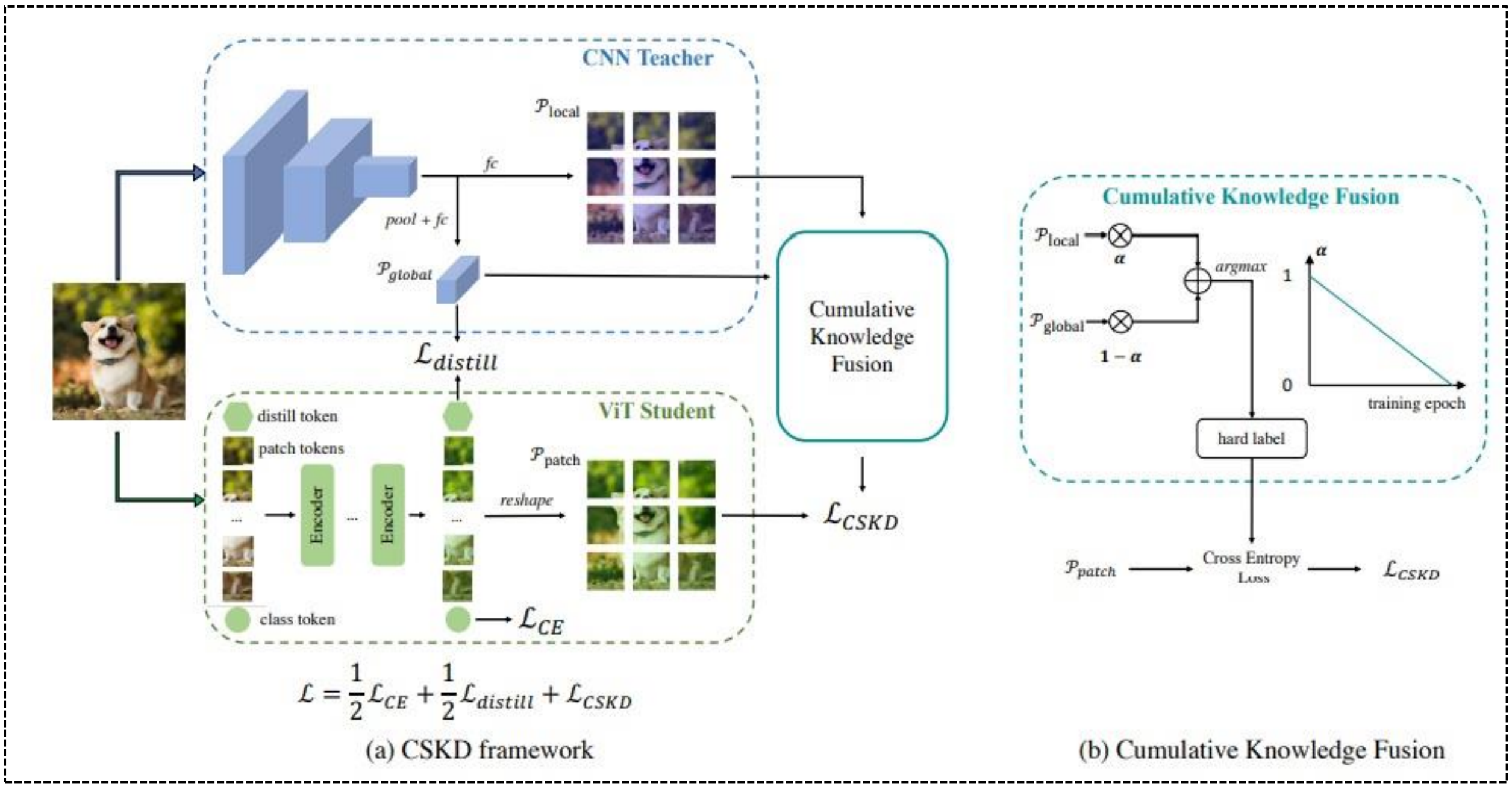

Figure 17. The Cumulative Spatial Knowledge Distillation (CSKD)

### 4.12) Self-Attention based Feature Matching with Spatial Transformers and Knowledge Distillation (SAM-Net)

The researchers in this paper [60] presented a novel matching approach that eliminates the necessity for traditional detectors, leveraging transformers and knowledge distillation. With the help of cross-attention layers within Transformers, the authors [60] developed SAM-Nets that dynamically adapt local features to local contexts as well as spatial relevance, thereby enhancing match quality significantly. Moreover, by employing PixLoc-style knowledge distillation techniques, the method effectively disregards dynamic object noise. The accuracy of localization is greatly improved by focusing on static objects. Through its achievement of state-of-the-art results, the corresponding approach [60] is demonstrated to be superior. We can get more details regrading ViTs and different knowledge distillation we can implement for compression in [61-65].
**Knowledge Distillation Approach:** Response based KD.

**KD loss function:** Kull back divergence distillation loss function + the coarse-level loss+ the fine-level loss, the flow estimation loss.

**Key Observations**

- The use of PixLoc-based knowledge distillation improves 2D feature matching and pose estimation by introducing a hierarchical attention mechanism.
- The method outperforms state-of-the-art methods in both indoor and outdoor datasets, as evidenced by high AUC scores.
- Scores high in visual localization benchmarks, with publicly available code and tests.

The overall framework of the proposed methodology can be illustrated in figure 18.

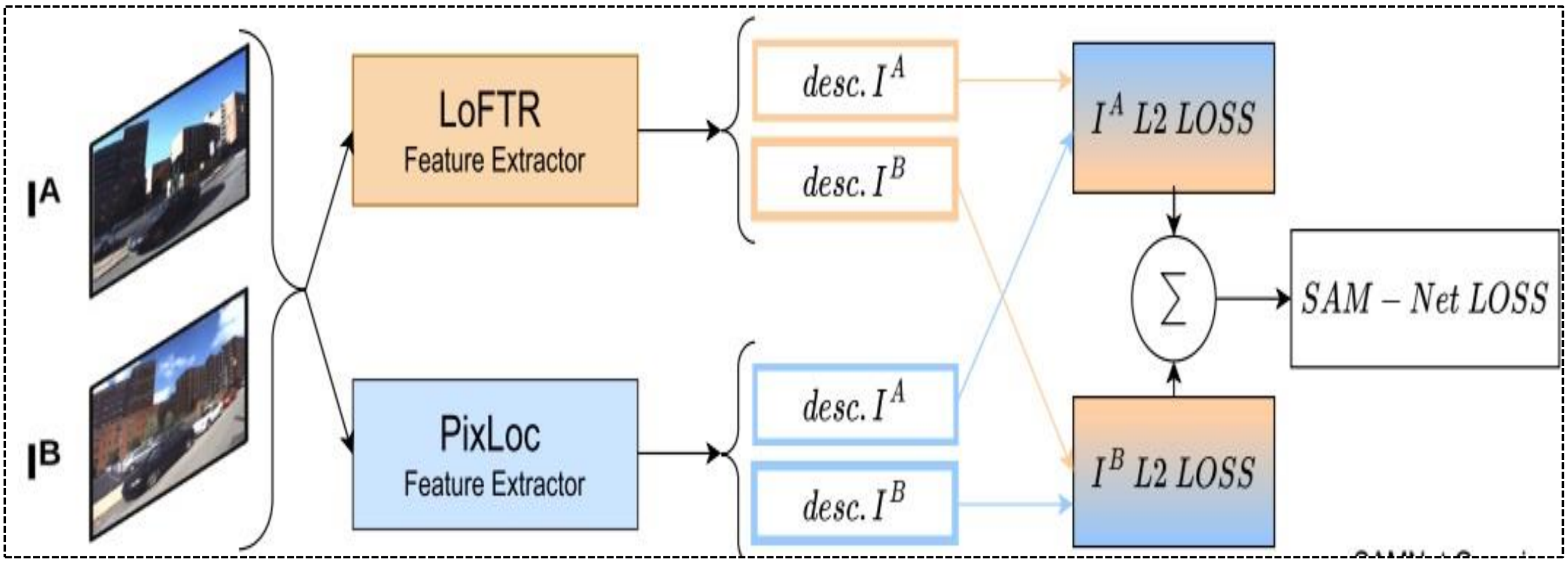

Figure 18. High-level Overview of KD with SAM Net framework

## 5). Comparative Analysis

After doing a comprehensive literature survey of various knowledge distillation methods in vision transformers, a summary of the methods is presented using different architectures for teachers and students under different knowledge distillation schemes in Table 1. Table 1 categorizes the different distillation methods under various distillation schemes and are clearly highlighted and colour coded in different colours. Also, in this section we would like to discuss various distillation loss functions using by above discussed methods and also mentioned in the table given below. The various loss functions used are discussed as:
As a first step, we will focus on the ImageNet dataset, a cornerstone of large-scale visual recognition. The ImageNet dataset, with its vast collection of higher-resolution images across a thousand categories, offers a unique opportunity, both in terms of challenges and opportunities. For ImageNet to be effective, sophisticated distillation strategies are necessary in order to transfer knowledge efficiently from teacher to student models, balancing accuracy and computational efficiency.
Here, we present a comprehensive summary of the results of applying various KD methods to different teacher-student configurations on the ImageNet dataset. This paper examines the trade-offs between improved top-level accuracy and a reduction in computational requirements, including FLOPs and model parameters, in relation to KD approaches, loss functions, and their impacts on model performance.

Table 1. ImageNet-1k Knowledge Distillation Approaches

| S.no | | Teacher → Student model | KD approach | KD loss function | Base Line Top 1 Test Accuracy (%) | Top-1 Accuracy (%) after KD | Base line Flops | Flops remained (%) after KD in Glops | Baseline Params | Params after KD | Reference |
|---|---|---|---|---|---|---|---|---|---|---|---|
| 1 | | **DeiT→DeiT-Tiny (Vit-Vit)** | Response-based KD+Skip-configuration +structured pruning (UVC) | L2 norm | 72.2 | 71.8 (-0.40) ↓ | 100 G | 53.0G | 6M | 2.32M | [59] |
| 2. | | **DeiT→DeiT-Tiny (Vit-Vit)** | Using only skip configuration | L2 norm | 72.2 | 68.72 (-3.48) ↓ | 100 G | 51.58G | 6M | 2.32M | [59] |
| 3 | | **DeiT→DeiT-Tiny (Vit-Vit)** | Structured pruning within the block | L2 norm | 72.2 | 70.50 (-1.75) ↓ | 100 G | 50.69G | 6M | 2.32M | [59] |
| 4 | | **DeiT→DeiT-Tiny (Vit-Vit)** | Without knowledge distillation | L2 norm | 72.2 | 69.34 ( -2.86) ↓ | 100 G | 51.23G | 6M | 2.32M | [59] |
| 5 | | **DeiT→DeiT-Tiny (Vit-Vit)** | UVC | L2 norm | 72.2 | 71.3 (-0.9) ↓ | 100 G | 49.23G | 6M | 2.32M | [59] |
| 6 | | **Diet→Diet-Small** | UVC | L2 norm | 79.8 | 78.82 (-0.98) ↓ | 100 G | 50.41G | 4.6M | 2.32M | [59] |
| 7 | | **T2T→ViT** | UVC | L2 norm | 81.5 | 78.9 (-2.6) ↓ | 100 G | 44.0 G | 4.8M | 2.11M | [59] |
| 8 | | **DeiT→DeiT- base** | UVC | L2 norm | 81.8 | 80.57(-1.23) ↓ | 100 G | 45.50 | 17.6M | 8.0M | [59] |
| 9. | | **Cait-XXs24→DeiT-Tiny** | Fine-grained manifold knowledge distillation method | Manifold distillation loss (intra image +inter image+ random sampled patch level distillation loss) | 78.5 | 75.5(-3.0) ↓ | N/A | N/A | N/A | N/A | [66] |
| 10 | | **Cait-S24→DeiT-Tiny** | Manifold knowledge distillation method | Manifold distillation loss (intra image +inter image+ random sampled patch level distillation loss) | 83.4 | 76.5 (-6.9) ↓ | N/A | N/A | N/A | N/A | [66] |

| | | | | | | | | | | |
|---|---|---|---|---|---|---|---|---|---|---|
| **11** | | **RegNetY-16GF→DeiT** | Convolution distillation from strong Convonet teacher. | Kull-back divergence loss function | 82.9 | 82.2 (-0.7) | N/A | N/A | 84M | 5M | [67] |
| **12** | | **RegNetY-600M(Conv)+RedNet26(inv)→CiT-Ti** | Convolution +involution KD approach | The weighted sum of two Kull back divergence loss functions | 74.0(convo + 76.0(invo.)) | 75.3 (-0.7) | N/A | N/A | 6M(convo)+ 9M(Invo) | 6M | [67] |
| **13** | | **ResNet-101→DeiT-Ti** | Data efficient early knowledge distillation with convolutional inductive biases (MHCA) | Cross entropy response+ mean square. distillation loss function | 77.37 | a) 72.2(train from scratch 74.6 (Distil on real images) (-2.4) b) 62.7 (Distilled on generated images with Deep Inversion (DI)) (-14.67) c) 70.1(ADI) (-7.37) d) 71.21(Distilled with DF-DearKD) (-6.37) | N/A | N/A | 45M | 5M | [68] |
| **14** | | **DeiT-B→DeiT-S** | Distillation through attention (via distillation token from a strong ConvNet classifier) | Kull back divergence +Cross entropy loss function | 81.8 | 84.0( -2.2) | N/A | N/A | 86M | 22M | [68],[69] |
| **15** | | **DeiT-B→DeiT-Ti** | Distillation through attention (via distillation token from a strong Convonet classifier) | Kull back divergence +Cross entropy loss function | 81.8 | 72.2 (8.8) | N/A | N/A | 86M | 5M | [68]-[69] |
| **16** | | | Pretraining distillation based upon | Cross entropy distillation loss function | 84.8 | 83.2 (-1.6) | N/A | N/A | 85.5M | 11.22 M | [70] |
| | | **Vit-L→Tiny-ViT** | saved sparse logits | | | | | | | | |
| **17** | | **Swin-L→Swin-T** | Pretraining distillation based upon saved sparse logits | Cross entropy distillation loss function | 84.2 | 81.2 (-3) ↓ | N/A | N/A | 84.4M | 28M | [70] |
| **18** | | **BEiT-L→Vit-Tiny** | Pretraining distillation based upon saved sparse logits | Cross entropy distillation loss function | 84.1 | 83.2(-0.9) ↓ | N/A | N/A | 84.4M | 11M | [70],[71], [72] |
| **19** | | **ViT-B→ResNet-50** | Cross-architecture knowledge distillation approach | Partially cross attention +Group-wise linear projector Multiview generator loss functions. | 82.7 | 78.34 (-4.36) ↓ | 55.4G | 4.1G | N/A | N/A | [73] |
| **20** | | **Swin-L→ResNet50** | Cross-architecture knowledge distillation approach | Partially cross attention +Group-wise linear projector Multiview generator loss functions. | 87.32 | 78.96 (-8.36) ↓ | 103.9G | 4.1G | N/A | N/A | [73], [74] |
| **21** | | **ViT-Base→ViT-Small** | New Supervised Masked Knowledge Distillation (SMKD) model for few-shot transformers that incorporates label information. | Contrastive Loss Function | | 89.57 | 80.0 (+9.57) ↑ | | 86M | 21M | [75], [76] |
| **22** | | **ViT-Base→Swin-Tiny** | (SMKD) | Contrastive Loss Function | | 89.57 | 89.96 | (+0.39) ↑ | 86M | 29.0M | [56] |
| **23** | | **RegNetY-16GF→CSKD-Ti** | Cumulative Spatial Knowledge Distillation (CSKD). | Cross Entropy loss+CSKD loss+ KLD distillation loss function | | 82.9 | 76.0 (-6.9) ↓ | | 86M | 5M | [57] |
| **24** | | **RegNetY-16GF→CSKD-Ts** | Cumulative Spatial Knowledge Distillation (CSKD). | Cross Entropy loss+CSKD loss+ KLD distillation loss function | | 82.9 | 82.3 (-0.6) ↓ | | 86M | 22M | [57] |
| **25** | | **RegNetY-16GF→ CSKD-Tb** | Cumulative Spatial Knowledge Distillation (CSKD). | Cross Entropy loss+CSKD loss+ KLD distillation loss function | | 82.9 | 83.8 (+0.9) ↑ | | 86M | 84M | [57] |

As a follow-up to our study of knowledge distillation (KD) techniques and their effects on the ImageNet dataset, we extend our analysis to the CIFAR-10 dataset. KD approaches present different challenges and opportunities when applying KD to CIFAR-10 because it is a smaller and less complex dataset than ImageNet. On the CIFAR-10 dataset, we present a comprehensive table detailing the results of various teacher-student model pairs, KD approaches, loss functions, and model efficiencies and accuracy. Results such as these highlight the effectiveness of KD techniques in improving model performance both in terms of accuracy and computing resources. The detailed analysis on cifar-10 dataset is highlighted in Table 2.

Table 2: Cifar-10 /100 Knowledge Distillation Approaches

| S.no | Teacher → Student model | KD approach | KD loss function | Base Line Top 1 Test Accuracy (%) | Top-1 Accuracy (%) after KD | Base line Flops | Flops remained (%) after KD in Glops | Baseline Params | Params after KD | Reference |
|---|---|---|---|---|---|---|---|---|---|---|
| 1 | **ResNet 110→ResNet 20** | Target aware transformer (TaT) approach | Patch group distillation loss function +Anchor point distillation loss function and Cross entropy (CE) | 74.31 | 71.70 (-2.61) ↓ | 0.255G | 0.041G | 1.7M | 0.27M | [46], [77] |
| 2 | **ResNet 56→ResNet 20** | Target aware transformer (TaT) approach | Patch group distillation loss function +Anchor point distillation loss function and Cross entropy (CE) | 72.00 | 70.06 (-2.06) ↓ | 0.2G | 0.041G | 0.85M | 0.27M | [46] – [77] |
| 3. | **ResNet-110→ResNet-32** | Target aware transformer (TaT) approach | Patch group distillation loss function +Anchor point distillation loss function and Cross entropy (CE) | 74.31 | 73.08(-1.23) ↓ | 0.25 M | 0.046G | 1.7M | 0.46M | [46]-[78] |
| 4 | **DeiT→DeiT-small** | Attention Probe distillation with Response based knowledge distillation approach. | Probe distillation +Kull back divergence loss function | 99.39 | 99.07(-0.32) ↓ | 1.01 G | 0.11G | 23.1M | 2.38M | [50],[79] |
| **5** | **DeiTX→DeiT - Tiny** | Attention Probe distillation with Response based knowledge distillation approach. | Probe distillation +Kull back divergence loss function | 99.04 | 87.66(-11.38) | 1.38 G | 0.15G | 153M | 5M | [50]-[79] |
| **6** | **DeiTX→ DeiT Tiny** <br> **Cifar-10 Unlabelled** | Attention Probe distillation with Response based knowledge distillation approach. | Probe distillation +Kull back divergence loss function. | 76.30 | 71.82 (-5.1) | 1.38 G | 0.15G | 21.3M | 2.38M | [50]-[80] |
| **7** | **DeiTX→ DeiT Tiny** <br> **Cifar-100 Unlabelled** | Attention Probe distillation with Response based knowledge distillation approach. | Probe distillation +Kull back divergence loss function | 96.65 | 93.95(-2.7) | 1.38 G | 0.15G | 153M | 2.38M | [80] |
| **8** | **ViT-B→ResNet-50** | Cross-architecture knowledge distillation approach | Partially cross attention +Group-wise linear projector Multiview generator loss functions. | 90.02 | 87.39 (-2.63) | 55.4 G | 4.1G | 86M | 25.4M | [67] , [81] |
| **9** | **Swin-L→ResNet50** | Cross-architecture knowledge distillation approach | Partially cross attention +Group-wise linear projector Multiview generator loss functions. | 87.32 | 76.28 (-11.04) | 103.9(G) | 4.1(G) | 197M | 25.4 | [67], [81] |
| **10** | **Swin-Tiny → EfficientNet-B0** | Cross-architecture knowledge distillation approach | Attention-based feature distillation + Adaptive temperature softmax loss. | 94.5 | 92.7 (-1.8) | 5.3G | 2.2G | 5.3M | 4.7M | [67], 83] |
| **11** | **Swin-L → MobileNetV2** | Cross-architecture knowledge distillation approach | Partially cross attention + Group-wise linear projector + Contrastive distillation loss function. | 93.50 | 88.34 (--5.16) | 103.9G | 0.3G | 197M | 6M | [67], [83], [84] |

In addition to the comprehensive analysis of the ImageNet /CIFAR-10/ Cifar-100 knowledge distillation (KD) techniques, we examined Pascal VOC data for object detection. The detection and localization of objects poses unique challenges compared with image classification. For accurate detection and localization of objects, KD techniques tailored to spatial and contextual understanding are required. With its diversity and complexity, Pascal VOC is an excellent benchmark for evaluating the effectiveness of KD approaches in improving object detection. To address the problem of detecting objects, we study systematically various configurations of teacher-student models, distillation strategies, and loss functions. Our research aims to shed light on how KD techniques may be used to improve detection accuracy as well as model efficiency, making advanced detection capabilities accessible with limited computing power. Based on the Pascal VOC dataset, Table 3 summarizes our findings in object detection using knowledge distillation.

Table 3: Pascal-VOC/ COCO dataset Knowledge Distillation Approaches

| **S.no** | **Teacher→Student model** | **Knowledge distillation (KD) approach** | **KD loss function** | **Top-1 Accuracy (%)** | **Top-1 Accuracy after KD** | **Total number of Flops** | **Flops remained (%) after KD** | **Baseline parameters** | **Student parameters** | **Reference** |
|---|---|---|---|---|---|---|---|---|---|---|
| 1. | **ResNet 101→ResNet 18** | TaT | Patch Anchor +(CE) | 78.43 | 75.76 (-2.67) ↓ | 7.8G | 1.9G | 44.5M | 11.2M | [78], [85], [86], [87] |
| 2.. | **ResNet-101→MobileneV2** | TaT | Patch Anchor +(CE) | 78.43 | 73.85 (-4.58) ↓ | 7.8G | 0.9G | 44.5M | 3.4M | [78], [86], [87], [88] |
| 3. | **Swin-small → Swin - Tiny + Mask R-CNN** | Manifold knowledge distillation method | Manifold distillation loss (intra image +inter image+ random sampled patch level distillation loss. | 83.2 | 82.2 (-1.00) ↓ | 365(G) | 272 G | 50M | 28M | [89], [90] |
| 4. | **Faster R-CNN → SSD** | Response-based KD | L2 norm + Cross-entropy | 76.4 | 74.2 (-2.00) ↓ | 100G | 11G | 138M | 21.8M | [91], [92] |
| 5. | **YOLOv3 → YOLOv3-Tiny** | Feature based Knowledge distillation | KDKL divergence + Objectness score loss | 72.6 | 70.3 (-2.3) ↓ | 65G | 6.8G | 63M | 8.3M | [91], [92], [93],[94] |
| 6 | **RetinaNet →MobileNetV2 + FPN** | Attention-guided KD | Focal loss + Attention map loss | 36.4 | 34.8 (-1.6) ↓ | 37G | 5.1G | 36M | 3.5M | [95], [96] |
| 7 | **R50DCN→R50** | Semantic Guided KD (feature distillation) | Mean square error loss + Cross Entropy loss | 41.42 | 45.04 (-3.62) ↓ | 5G | 4.1G | 30M | 25.6M | [97], [98] |
| 8 | **ViT-L→Faster-RCNN** | Structured KD | Classification loss + Regression loss | 51.5 | 50.3 ( -1.2) ↓ | 55.0G | 28.1G | 86.0M | 30.57M | [99], [100] |
| 9. | **ViT-B→Faster RCNN** | Structured KD | Classification loss + Regression loss | 55.6 | 52.3 ( -1.2) ↓ | 49.3G | 28.1G | 56M | 30.57M | [99], [100] |
| 10 | **ViT-S→Faster RCNN** | Structured KD | Classification loss + Regression loss | 48.98 | 49.08 (-0.1) ↓ | 48.0G | 28.1G | 40M | 30.57M | [99], [100] |

As a result of a comprehensive analysis of multiple datasets, we found that knowledge distillation (KD) techniques are applied consistently in Visual Transformers (ViTs). All KD methods we explored contribute to the overarching goal of model efficiency positively. Furthermore, these techniques enable the development of models that use fewer computational resources, making them more suitable for environments with limited resources.

## 5) Trade-off between Computational Efficiency and Performance in Vision Transformers through Knowledge Distillation"

This comprehensive study highlights an important trade-off between computational resources and model performance when optimizing Vision Transformers (ViTs) through Knowledge Distillation (KD). As evidenced by methods like pretraining distillation and structured pruning, which are able to significantly reduce model size and computational demands, KD techniques manage to achieve this balance with minimal impact on model accuracy while reducing model size and computational requirements. As an example, models like Tiny ViT can achieve considerable parameter reductions with no significant performance loss, demonstrating their utility for resource-constrained deployments. With these approaches, models can perform exceptionally well on targeted tasks such as image classification and facial recognition while reducing parameters by up to 75% and decreasing inference time by up to 65%. In environments where resources are at a premium, optimized ViTs offer a path to utilizing advanced AI capabilities on mobile and embedded devices without significantly sacrificing performance, due to their balance between efficiency, size, and accuracy. Consequently, it prompts a careful assessment of the application requirements, guiding the selection of KD methods that meet the computational constraints while satisfying performance requirements. It is important to strike the right balance between the computational budget and the requirement for high accuracy in this field to achieve optimal results. The downside of this efficiency is that sometimes accuracy is sacrificed at the expense of efficiency, which is pivotal for applications that require precision. This trade-off highlights a fundamental challenge in the field: finding a balance between the computational budget and high accuracy requirements. Consequently, it guides selection of KD methods that meet computational constraints while meeting performance expectations by carefully evaluating application-specific requirements. Therefore, our framework allows us to navigate the trade-off between performance and resource consumption when deploying lightweight yet effective Vision Transformers.

## 6) Conclusion and Limitations of the Study

Visual Transformers have gained considerable attention since they demonstrated their effectiveness in CV tasks and undercut CNNs' dominance in the CV domain. This paper discussed ViT in detail and the open challenges faced by ViT in computer vision tasks. Also, the paper provided a deep insight into aggressive compression techniques to make these large-scale models computationally and resource-efficient. The ViTs have achieved significant progress in multiple benchmarks and obtained promising results comparable to or even better than the SOTA CNN methods. Some key technologies for ViTs still need to be improved to deal with complex challenges faced by ViT in CV tasks. One of the strengths of ViT models is their ability to scale to high parametric complexity. However, this remarkable property allows the Training of extremely large models but incurs high training and inference costs. These models require huge computational resources, which are expensive, and come up with a huge price. Furthermore, real-world applications of these large-scale models require aggressive compression (e.g., distillation). Future work will focus on compressing such large-scale models to make them feasible for resource-constrained environments and to reduce the computational and latency requirements.

## 7) Future directions

Following presents the future directions pertaining to KD in ViTs;

- Even though self-attention allows us to model full-image contextual information, it incurs memory and computation cost. To capture local and global contextual pixel information, the attention mechanism comes up with huge time complexity of $O(N^2)$, where N represents several input features maps. A criss-cross attention module and knowledge distillation must be employed to generate a sparse attention map on the cross path to reduce this computational burden. It uses 11 times less GPU memory than non-local blocks and has a complexity of O (2 √ N).

- Despite providing competitive results as a stand-alone computational primitive, the best results are obtained when combined with convolutional neural networks. Attention augmentation can be explored and used, leading to systematic performance gains for image classification and object detection across different architectures.
- It remains an unsolved problem to visualize and interpret Transformers, requiring methods for obtaining spatially precise activation-specific visualizations. As progress is made in this direction, the Transformer models can be better understood, as can diagnose erroneous behaviours and biases in the decision-making process. It can also facilitate the design of novel architectures that will allow us to avoid biases.
- In the case of large feature maps, above discussed methods are likely to become intractable since these methods calculate correlations between feature spatial locations. So, remains an open challenge to combat when finetuning at higher resolutions.
- From the above study, there are three main reasons scaling up performs better: (a) large models (with more parameters) can benefit from more training data, and smaller models plateau quickly. They can't benefit from more training samples. Therefore, large-scale models can further improve their representation learning abilities.
- The recent transformer model known as Restormer argues that if Transformer architecture is scaled up, it should be a broader or deeper design. While broad models can reduce computational time via parallelization, deeper architectures deliver better performance.
- Relative position encodings are noted as superior to absolute position encodings in many of the above-discussed studies. However, it is still being determined what causes this difference. Therefore, the benefits and drawbacks of different position encoding methods need to be systematically studied and understood.